%% file: paper.tex
\documentclass[runningheads]{llncs}
\usepackage{graphicx}

\usepackage{tikz}
\usepackage{comment} 
\usepackage{amsmath,amssymb} 
\usepackage{color}

\usepackage[width=122mm,left=12mm,paperwidth=146mm,height=193mm,top=12mm,paperheight=217mm]{geometry}

\usepackage{microtype}
\usepackage{multirow}
\usepackage{kotex}

\input{header}

\begin{document}
\pagestyle{headings}
\mainmatter
\def\ECCVSubNumber{1644}  

\title{Neural Geometric Parser for Single Image Camera Calibration\thanks{Corresponding author: Junho Kim (junho@kookmin.ac.kr)}}

\author{Jinwoo Lee\inst{1} \and
  Minhyuk Sung\inst{2} \and
  Hyunjoon Lee\inst{3} \and
  Junho Kim\inst{1}}
%
\authorrunning{J. Lee et al.}
%
\institute{Kookmin University \, \email{\{jinwoolee,junho\}@kookmin.ac.kr} \and
  Adobe Research \, \email{msung@adobe.com} \and
  Intel Korea \, \email{eldercrow@gmail.com}
}
\institute{
\textsuperscript{1}Kookmin University \quad  \textsuperscript{2}Adobe Research \quad  \textsuperscript{3}Intel
}

\maketitle

\graphicspath{{./fig/}}

\input{sections/abstract}
\input{sections/introduction}

\input{sections/related_work}

\input{sections/framework}

\input{sections/experiments}
\input{sections/conclusion}

\bigbreak
{\footnotesize
\parahead{Acknowledgements}
This research was supported by the National Research Foundation of Korea (NRF) funded by the Ministry of Education (2017R1D1A1B03034907).
}

\clearpage
%
%
\bibliographystyle{splncs04}
\bibliography{references}

\clearpage

\renewcommand{\thesection}{A}
\setcounter{table}{0}
\renewcommand{\thetable}{A\arabic{table}}
\setcounter{figure}{0}
\renewcommand{\thefigure}{A\arabic{figure}}

\newif\ifpaper
\papertrue

\section*{Appendix}
\input{sections/appendix}

\end{document}

\end{document}

%% file: header.tex
\usepackage{amsfonts,amsmath,amssymb}
\usepackage{booktabs}
\usepackage{colortbl}
\usepackage{comment}
\usepackage{dsfont}
\usepackage{graphicx}
\usepackage{makecell}
\usepackage{multirow}
\usepackage{rotating}
\usepackage{scalerel}
\usepackage[group-separator={,}]{siunitx}
\usepackage{tabularx}
\usepackage{wrapfig}
\usepackage{xspace}
\usepackage{capt-of}

\newcolumntype{L}{>{\raggedright\arraybackslash}X}
\newcolumntype{C}{>{\centering\arraybackslash}X}

\newcommand{\Eq}[1]  {Eq.\ (\ref{eq:#1})}
\newcommand{\Eqs}[1] {Eqs.\ (\ref{eq:#1})}
\newcommand{\Fig}[1] {Fig.\ \ref{fig:#1}}

\newcommand{\Tbl}[1]  {Table~\ref{tbl:#1}}

\newcommand{\Sec}[1] {Sec.\ \ref{sec:#1}}

\makeatletter
\DeclareRobustCommand\onedot{\futurelet\@let@token\@onedot}
\def\@onedot{\ifx\@let@token.\else.\null\fi\xspace}

\def\etal{\emph{et al}\onedot}
\makeatother

\newcommand{\parahead}[1]{\noindent\textbf{#1}.\ }

\definecolor{Goldenrod}{rgb}{0.85, 0.65, 0.13}
\definecolor{Orchid}{rgb}{0.85, 0.44, 0.84}
\definecolor{LimeGreen}{rgb}{0.2, 0.8, 0.2}
\definecolor{OliveGreen}{rgb}{0.25, 0.49, 0.19}
\definecolor{SkyBlue}{rgb}{0.53, 0.81, 0.92}
\definecolor{LightCyan}{rgb}{0.88,1,1}
\definecolor{LightGoldenrod}{rgb}{1.00, 0.93, 0.55}

%% file: sections/abstract.tex

\begin{abstract}
We propose a neural geometric parser learning single image camera calibration for man-made scenes. Unlike previous neural approaches that rely only on semantic cues obtained from neural networks, our approach considers both semantic and geometric cues, resulting in significant accuracy improvement. The proposed framework consists of two networks. Using line segments of an image as geometric cues, the first network estimates the zenith vanishing point and generates several candidates consisting of the camera rotation and focal length. The second network evaluates each candidate based on the given image and the geometric cues, where prior knowledge of man-made scenes is used for the evaluation. With the supervision of datasets consisting of the horizontal line and focal length of the images, our networks can be trained to estimate the same camera parameters. Based on the Manhattan world assumption, we can further estimate the camera rotation and focal length in a weakly supervised manner. The experimental results reveal that the performance of our neural approach is significantly higher than that of existing  state-of-the-art camera calibration techniques for single images of indoor and outdoor scenes.
\keywords{Single image camera calibration, Neural geometric parser, Horizon line, Focal length, Vanishing Points, Man-made scenes}
\end{abstract}

%% file: sections/introduction.tex
	
\section{Introduction}
\label{sec:introduction}

\begin{figure*}[t!]
\centering
\begin{tabular}{cccc}
\includegraphics[height=0.2\linewidth]{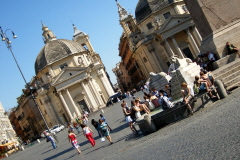} &
\includegraphics[height=0.2\linewidth]{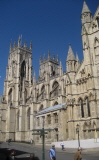} &
\includegraphics[height=0.2\linewidth]{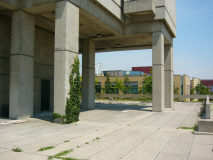} &
\includegraphics[height=0.2\linewidth]{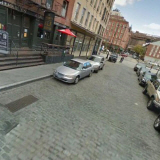} \\
\includegraphics[height=0.2\linewidth]{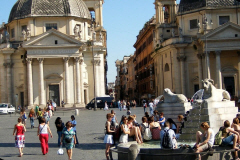} &
\includegraphics[height=0.2\linewidth]{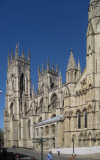} &
\includegraphics[height=0.2\linewidth]{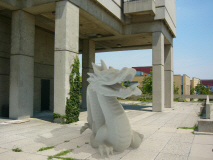} &
\includegraphics[height=0.2\linewidth]{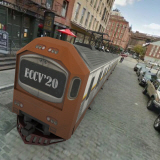} \\
(a) & (b) & (c) & (d) \\
\end{tabular}
\caption{Applications of the proposed framework: (a) image rotation correction, (b) perspective control, and virtual object insertions with respect to the (c) horizon and (d) VPs; before (top) and after (bottom).}
\label{fig:teaser}
\end{figure*}

This paper deals with the problem of inferring camera calibration parameters from a single image. It is used in various applications of computer vision and graphics, including image rotation correction~\cite{Fischer:2015}, perspective control~\cite{Lee:2014}, camera rotation estimation~\cite{Zhou:2019}, metrology~\cite{Criminisi:2000}, and 3D vision~\cite{MVG,Ma:3DV}. Due to its importance, single image camera calibration has been revisited in various ways.

Conventional approaches focus on reasoning vanishing points (VPs) in images by assembling geometric cues in the images. Most methods find straight line segments in the images using classic image processing techniques~\cite{Gioi:2010,Akinlar:2011} and then estimate the VPs by carefully selecting parallel or orthogonal segments in the 3D scene as geometric cues~\cite{Lee:2014}. In practice, however, line segments found in images contain a large amount of noisy data, and it is therefore important to carefully select an inlier set of line segments for the robust detection of VPs~\cite{RANSAC:1981,Tardif:2009}. Because the accuracy of the inlier set is an important performance indicator, the elapsed time may exponentially increase if stricter criteria are applied to draw the inlier set.

Recently, several studies have proposed estimating camera intrinsic parameters using semantic cues obtained from deep neural networks. It has been investigated~\cite{Workman:2015,Workman:2016,Hold-Geoffroy:2018} that well-known backbone networks, such as ResNet~\cite{He:2016} and U-Net~\cite{Ronneberger:2015}, can be used to estimate the focal length or horizon line of an image without significant modifications of the networks. In these approaches, however, it is difficult to explain which geometric interpretation inside of the networks infers certain camera parameters. In several studies~\cite{Zhou:2019,Xian:2019}, neural networks were designed to infer geometric structures; however, they required a new convolution operator~\cite{Zhou:2019} or 3D supervision datasets~\cite{Xian:2019}.

In this paper, we propose a novel framework for single image camera calibration that combines the advantages of both conventional and neural approaches. The basic idea is for our network to leverage line segments to reason camera parameters. We specifically focus on calibrating camera parameters from a single image of a man-made scene. By training with image datasets annotated with horizon lines and focal lengths, our network infers pitch, roll, and focal lengths (3DoF) and can further estimate camera rotations and focal lengths through three VPs (4DoF).

\begin{figure*}[t!]
\includegraphics[width=1\linewidth]{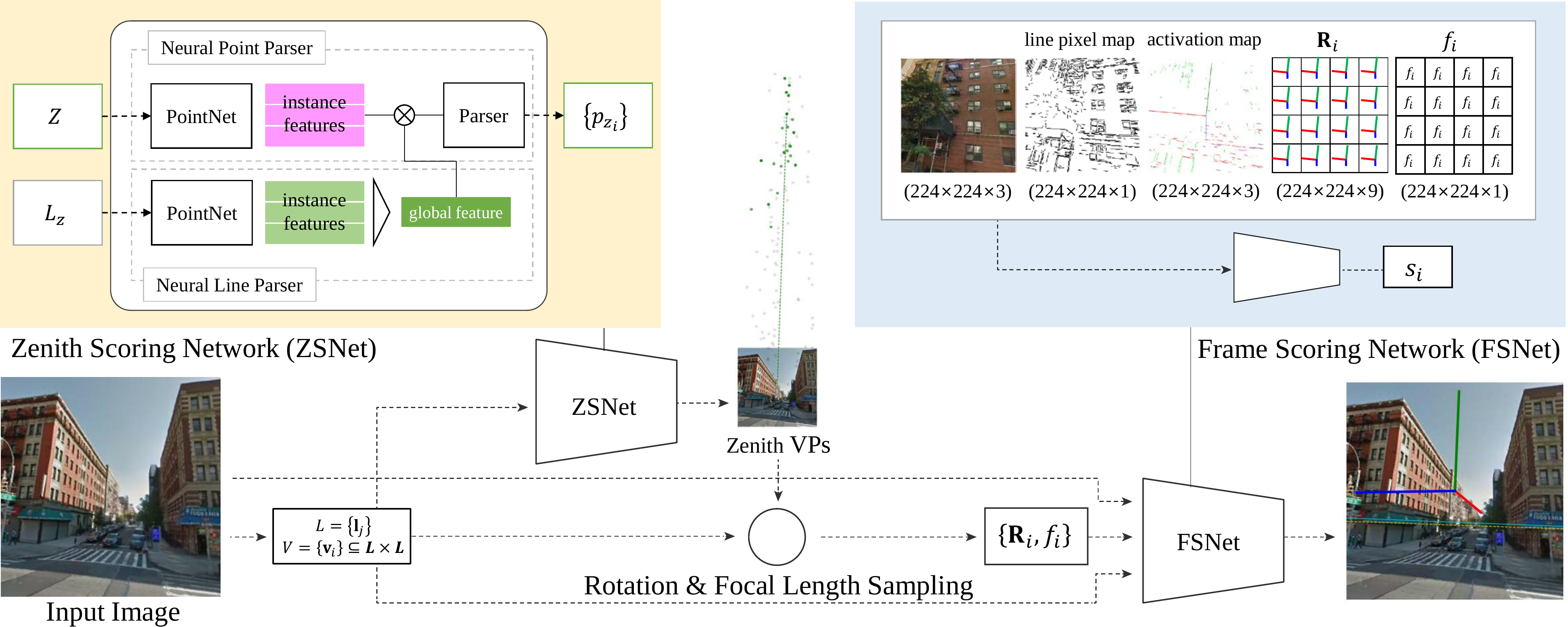} 
\caption{Overview of the proposed neural geometric parser.}    
\label{fig:overview}
\end{figure*}

The proposed framework consists of two networks. The first network, the Zenith Scoring Network (ZSNet), takes line segments detected from the input image and deduces reliable candidates of parallel world lines along the zenith VP. Then, from the lines directed at the zenith VP, we generate candidate pairs consisting of a camera rotation and a focal length as inputs of the following step. The second network, the Frame Scoring Network (FSNet), evaluates the score of its input in conjunction with the given image and line segment information. Here, geometric cues from the line segments are used as prior knowledge about the man-made scenes in our network training. This allows us to obtain significant improvement over previous neural methods that only use semantic cues \cite{Workman:2016,Hold-Geoffroy:2018}. Furthermore, it is possible to estimate camera rotation and focal length in a \emph{weakly supervised} manner based on the Manhattan world assumption, as we reason camera parameters with pairs consisting of a camera rotation and a focal length. It should be noted that the ground truth for our supervisions is readily available with Google Street View~\cite{GSVI} or with consumer-level devices possessing a camera and an inertial measurement unit (IMU) sensor, in contrast to the method in \cite{Xian:2019}, which requires 3D supervision datasets.

%% file: sections/related_work.tex

\section{Related Work}
\label{sec:related_work}

Projective geometry~\cite{MVG,Ma:3DV} has historically stemmed from the study on imaging of perspective distortions occurring in the human eyes when one observes man-made architectural scenes~\cite{Alberti:1436}. In this regard, conventional methods of single image camera calibration~\cite{Coughlan:1999,Kosecka:2002,Schindler:2004,YUD:2008,ECD:2012,Wildenauer:2012,Xu:2013,Lee:2014} involve extracting line segments from an image, inferring the combinations of world parallel or orthogonal lines, identifying two or more VPs, and finally estimating the rotation and focal length of the camera. LSD~\cite{Gioi:2010} or  EDLine~\cite{Akinlar:2011} were commonly used as effective line segment detectors, and RANSAC~\cite{RANSAC:1981} or J-Linkage~\cite{Tardif:2009} were adopted to identify VPs describing as many of the extracted line segments as possible. Lee~\etal~\cite{Lee:2014} proposed robust estimation of camera parameters and automatic adjustment of camera poses to achieve perspective control. Zhai~\etal~\cite{Zhai:2016} analyzed the global image context with a neural network to estimate the probability field in which the horizon line was formed. In their work, VPs were inferred with geometric optimization, in which horizontal VPs were placed on the estimated horizon line. Simon~\etal~\cite{Simon:2018} achieved better performance than Zhai~\etal~\cite{Zhai:2016} by inferring the zenith VP with a geometric algorithm and carefully selecting a line segment orthogonal to the zenith VP to identify the horizon line. Li~\etal~\cite{Li:2019} proposed a quasi-optimal algorithm to infer VPs from annotated line segments.

Recently, neural approaches have been actively studied to infer camera parameters from a single image using semantic cues learned by convolutional neural networks. Workman~\etal proposed DeepFocal~\cite{Workman:2015} for estimating focal lengths and DeepHorizon~\cite{Workman:2016} for estimating horizon lines using semantic analyses of images with neural networks. Hold-Geoffroy~\etal~\cite{Hold-Geoffroy:2018} trained a neural classifier that jointly estimates focal lengths and horizons. They demonstrated that their joint estimation leads to more accurate results than those produced by independent estimations~\cite{Workman:2015,Workman:2016}. Although they visualized how the convolution filters react near the edges (through the method proposed by Zeiler and Fergus~\cite{Zeiler:2014}), it is difficult to intuitively understand how the horizon line is geometrically determined through the network. Therefore, \cite{Workman:2015,Workman:2016,Hold-Geoffroy:2018} have a common limitation that it is non-trivial to estimate VPs from the network inference results. 
Zhou~\etal~\cite{Zhou:2019} proposed NeurVPS that infers VPs with conic convolutions for a given image. However, NeurVPS~\cite{Zhou:2019} assumes normalized focal lengths and does not estimate focal lengths.

Inspired by UprightNet~\cite{Xian:2019}, which takes geometric cues into account, we propose a neural network that learns camera parameters by leveraging line segments. Our method can be compared to Lee~\etal~\cite{Lee:2014} and Zhai~\etal~\cite{Zhai:2016}, where line segments are used to infer the camera rotation and focal length. However, in our proposed method, the entire process is designed with neural networks. Similar to Workman~\etal~\cite{Workman:2016} and  Hold-Geoffroy~\etal~\cite{Hold-Geoffroy:2018}, we utilize semantic cues from neural networks but our network training differs in that line segments are utilized as prior knowledge about man-made scenes. Unlike UprightNet~\cite{Hold-Geoffroy:2018}, which requires the supervisions of depth and normal maps for learning roll/pitch (2DoF), the proposed method learns the horizon and focal length (3DoF) with supervised learning and the camera rotation and focal length (4DoF) with weakly supervised learning.

The relationship between our proposed method and the latest neural RANSACs \cite{Dsac:2017,Brachmann:2019,Kluger:2020} is described below. Our ZSNet is related to neural-guided RANSAC \cite{Brachmann:2019} in that it updates the line features with backpropagation when learning to sample zenith VP candidates. In addition, our FSNet is related to DSAC \cite{Dsac:2017} in that it evaluates each input pair consisting of a camera rotation and focal length based on the hypothesis on man-made scenes. Our work differs from CONSAC \cite{Kluger:2020}, which requires the supervision of all VPs, as we focus on learning single image camera calibrations from the supervision of horizons and focal lengths.

%% file: sections/framework.tex

\section{Neural Geometric Parser for Camera Calibration}
\label{sec:framework}

From a given input image, our network estimates up to four camera intrinsic and extrinsic parameters; the focal length $f$ and three camera rotation angles $\psi$, $\theta$, $\phi$.
Then, a 3D point $(P_x, P_y, P_z)^T$ in the world coordinate is projected onto the image plane as follows:
\begin{equation}
    \left[
        \begin{array}{c}
            p_x \\ p_y \\ p_w
        \end{array}
    \right] =
    \left( \mathbf{KR} \right) \left[
        \begin{array}{c}
            P_x \\ P_y \\ P_z
        \end{array}
    \right] \mathrm{,~where~} 
    \mathbf{K} = \left[
		\begin{array}{ccc}
			f & 0 & c_u \\
			0 & f & c_v \\
			0 & 0 & 1
		\end{array} \right]
	\mathrm{~and~}
	\mathbf{R} = \mathbf{R}_\psi\mathbf{R}_\theta\mathbf{R}_\phi,
\label{eq:camera_projection}
\end{equation}
where $(p_x, p_y, p_w)^{T}$ represents the mapped point in the image space, and $\mathbf{R}_\psi$, $\mathbf{R}_\theta$, $\mathbf{R}_\phi$ represent the rotation matrices along $x$-, $y$-, and $z$-axes, with rotation angles $\psi$, $\theta$, $\phi$, respectively. The principal point is assumed to be on the image center such that $c_u = W / 2$ and $c_v = H / 2$, where $W$ and $H$ represent the width and height of the image, respectively.

Under the Manhattan world assumption, calibration can be done once we obtain the \textit{Manhattan directions}, which are three VPs corresponding to $x$-, $y$-, and $z$-directions in 3D \cite{Coughlan:1999}. 
In \Sec{zsnet}, we describe how to extract a set of candidate VPs along the zenith direction. Then, in \Sec{fsnet}, we present our weakly supervised method for estimating all three directions and calibrating the camera parameters.

We use LSD~\cite{Gioi:2010} as a line segment detector in our framework. A line segment is represented by a pair of points in the image space. Before estimating the focal length in \Sec{fsnet}, we assume that each image is transformed into a pseudo camera space as $\mathbf{p} = \mathbf{K}_p^{-1} \left( p_x, p_y, p_w \right)^T$, where $\mathbf{K}_p$ represents a pseudo camera intrinsic matrix of $\mathbf{K}$, built by assuming $f$ as $\min(W, H) / 2$.

\subsection{Zenith Scoring Network (ZSNet)}
\label{sec:zsnet}

We first explain our ZSNet, which is used to estimate the zenith VP (see \Fig{overview} top-left). Instead of searching for a single zenith VP, we extract a set of candidates that are sufficiently close to the ground truth.

Similar to PointNet \cite{Qi:2017}, ZSNet takes sets of unordered vectors in 2D homogeneous coordinates - line equations and VPs - as inputs. Given a line segment, a line equation $\mathbf{l}$ can be computed as a cross product of its two endpoints:
\begin{equation}
	\mathbf{l} = \left[\mathbf{p}_0\right]_{\times} \mathbf{p}_1,
\label{eq:line_eq}
\end{equation}
where $[\cdot]_{\times}$ represents a skew-symmetric matrix of a vector.
A candidate VP $\mathbf{v}$ can then be computed as an intersection point of the two lines:
\begin{equation}
	\mathbf{v} = \left[\mathbf{l}_0\right]_{\times} \mathbf{l}_1.
\label{eq:vp_eq}
\end{equation}

Motivated by~\cite{Simon:2018}, we sample a set of line equations roughly directed to the zenith $L_z = \left\{ \mathbf{l}_0, \ldots, \mathbf{l}_{|L_z|} \right\}$ from the line segments, using the following equation:
\begin{equation}
    \left| \tan^{-1}\left( -\frac{a}{b} \right) \right| > \tan^{-1}\left( \delta_z \right),
\label{eq:angle_threshold}
\end{equation}
where $\mathbf{l} = (a, b, c)^{T}$ represents a line equation as in \Eq{line_eq} and the angle threshold $\delta_z$ is set to $67.5^\circ$ as recommended in~\cite{Simon:2018}. Then, we randomly select pairs of line segments from $L_z$ and compute their intersection points as in \Eq{vp_eq} to extract a set of zenith VP candidates $Z = \left\{ \mathbf{z}_0, \ldots, \mathbf{z}_{|Z|} \right\}$.
Finally, we feed $L_z$ and $Z$ to ZSNet. We set both the number of samples, $|L_z|$ and $|Z|$, \num{256} in the experiments.

The goal of our ZSNet is to score each zenith candidate in $Z$; 1 if a candidate is sufficient close to the ground truth zenith, and 0 otherwise. \Fig{overview} top-left shows the architecture of our ZSNet. 

In the original PointNet \cite{Qi:2017}, each point is processed independently, except for transformer blocks, to generate point-wise features. A global max pooling layer is then applied to aggregate all the features and generate a global feature. The global feature is concatenated to each point-wise feature, followed by several neural network blocks to classify/score each point.

In our network, we also feed the set of zenith candidates $Z$ to the network, except that we do not compute the global feature from $Z$. Instead, we use another network, feeding the set of line equations $L_z$, to extract the global feature of $L_z$ that is then concatenated with each point-wise feature of $Z$ (\Fig{overview}, top-left).

Let $h_z(\mathbf{z}_i)$ be a point-wise feature of the point $\mathbf{z}_i$ in $Z$, where $h_z(\cdot)$ represents a PointNet feature extractor. Similarly, let $h_l(L_z) = \left\{ h_l(\mathbf{l}_0), \ldots, h_l(\mathbf{l}_{|L_z|}) \right\}$ be a set of features of $L_z$. A global feature $\mathbf{g}_l$ of $h_l(L_z)$ is computed via a global max-pooling operation ($\mathrm{gpool}$), and is concatenated to $h_z(\mathbf{z}_i)$ as follows:
\begin{align}
    \mathbf{g}_l &= \mathrm{gpool} \left( h_l(L_z) \right) 
    \label{eq:line_feat} \\
    h'_z(\mathbf{z}_i) &= \mathbf{g}_l \otimes h_z(\mathbf{z}_i),
    \label{eq:feat_concat}
\end{align}
where $\otimes$ represents the concatenation operation.

Finally, the concatenated features are fed into a scoring network computing $[0, 1]$ scores such that:
\begin{equation}
p_{z_i} = \mathrm{sigmoid}\left( s_z(h'_z(\mathbf{z}_i)) \right),
\label{eq:zenith_pred}
\end{equation}
where $p_{z_i}$ represents the computed scores of each zenith candidate. The network $s_z(\cdot)$ in \Eq{zenith_pred} consists of multiple MLP layers, similar to the latter part of the PointNet \cite{Qi:2017} segmentation architecture.

To train the network, we assign a ground truth label $y_i$ to each zenith candidate $\mathbf{z}_i$ using the following equation:
\begin{equation}
    y_i = \left\{
        \begin{array}{ll}
            1 & \mathrm{~if~} \mathrm{cossim}(\mathbf{z}_i, \mathbf{z}_{gt}) > \cos(\delta_p) \\
            0 & \mathrm{~if~} \mathrm{cossim}(\mathbf{z}_i, \mathbf{z}_{gt}) < \cos(\delta_n)
        \end{array} \right.,
\label{eq:zenith_label}
\end{equation}
where $\mathrm{cossim}(x,y) = \frac{\left|x \cdot y \right|}{\|x\|\|y\|}$ and $\mathbf{z}_{gt}$ represents the ground truth zenith. The two angle thresholds $\delta_p$ and $\delta_n$ are empirically selected as $2^\circ$ and $5^\circ$, respectively, from our experiments. The zenith candidates each of which $y_i$ is undefined are not used in the training. The cross entropy loss is used to train the network as follows:
\begin{equation}
\mathcal{L}_{cls} = \frac{1}{N} \sum_i^N -y_i \log(p_{z_i}).  
\end{equation}

To better train our ZSNet we use another loss in addition to the cross entropy. Specifically, we constrain the weighted average of zenith candidates close to the ground truth, where estimated scores $p_{z_i}$ are used as weights. To average the zenith candidates, which represent vertical directions of the scene, we use structure tensors of $l_2$ normalized 2D homogeneous points. Given a 2D homogeneous point $\mathbf{v} = (v_x, v_y, v_w)^{T}$, a structure tensor of the normalized point is computed as follows: 
\begin{equation}
    \mathrm{ST}(\mathbf{v}) = \frac{1}{\left( v_x^2 + v_y^2 + v_w^2 \right)}
    \left[
    \begin{array}{ccc}
        v_x^2 & v_x v_y & v_x v_w \\
        v_x v_y & v_y^2 & v_y v_w \\
        v_x v_w & v_y v_w & v_w^2
    \end{array}
    \right],
\label{eq:structure_tensor}
\end{equation}
The following loss is used in our network:
\begin{equation}
    \mathcal{L}_{loc} = \left\Vert \mathrm{ST}(\mathbf{z}_{gt}) -
            \overline{\mathrm{ST}}(\mathbf{z}) \right\Vert_F, ~~
    \overline{\mathrm{ST}}(\mathbf{z}) = 
            \frac{\sum_i p_{z_i} \mathrm{ST}(\mathbf{z}_i)}{\sum_i p_{z_i}}
\label{eq:zenith_loc_loss}
\end{equation}
where $\Vert \cdot \Vert_F$ represents the Frobenius norm. Finally, we select zenith candidates whose scores $p_{z_i}$ are larger than $\delta_c$ to the set of zenith candidates, as:
\begin{equation}
    Z_c = \left\{ \mathbf{z}_i ~|~ p_{z_i} > \delta_c \right\},
    \label{eq:zenith_score}
\end{equation}
where $\delta_c = 0.5$ in our experiments. The set $Z_c$ is then used in our FSNet.

\subsection{Frame Scoring Network (FSNet)}
\label{sec:fsnet}

\begin{figure}[t!]
    \centering
    \begin{tabular}{cccc}
    \includegraphics[width=0.24\linewidth]{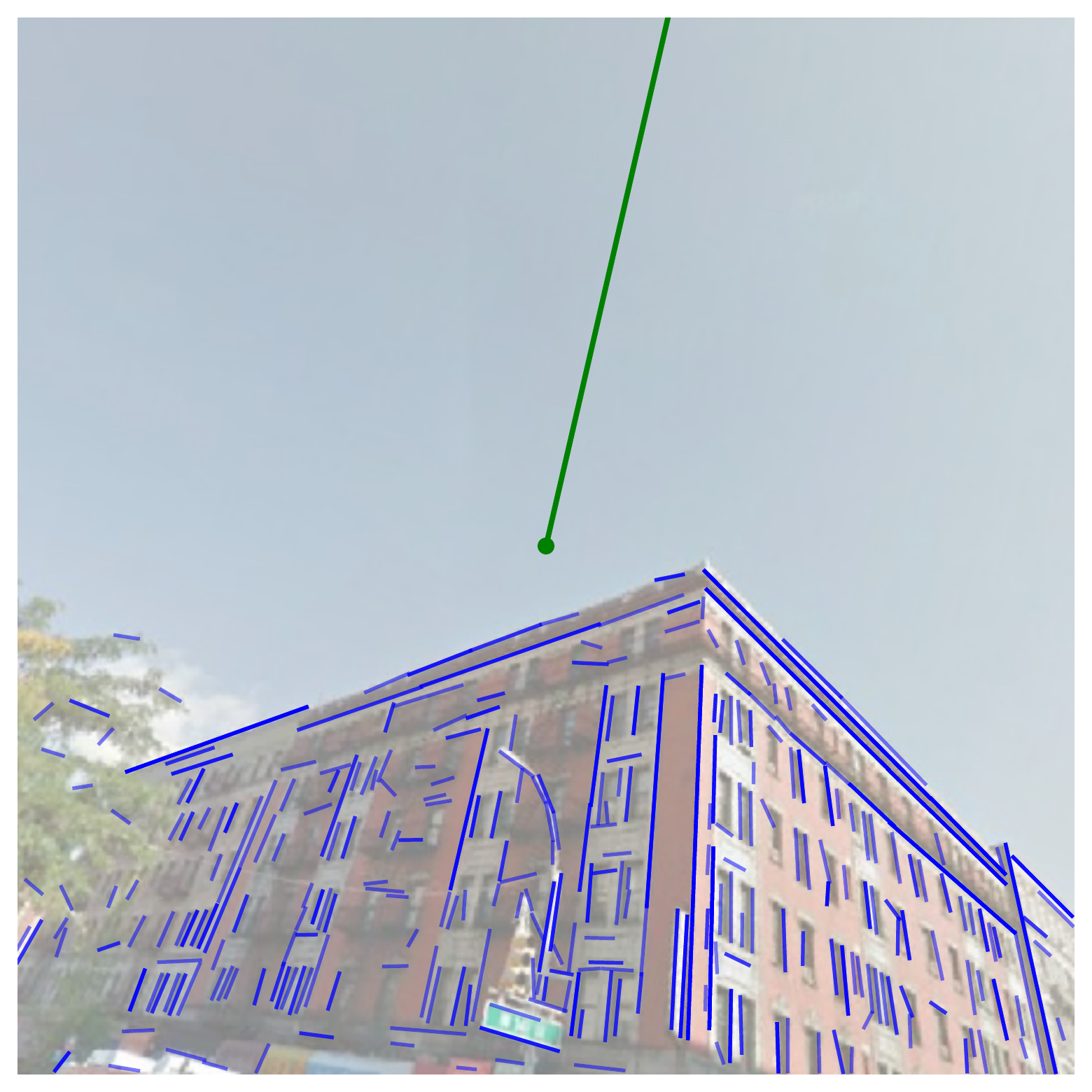} &
    \includegraphics[width=0.24\linewidth]{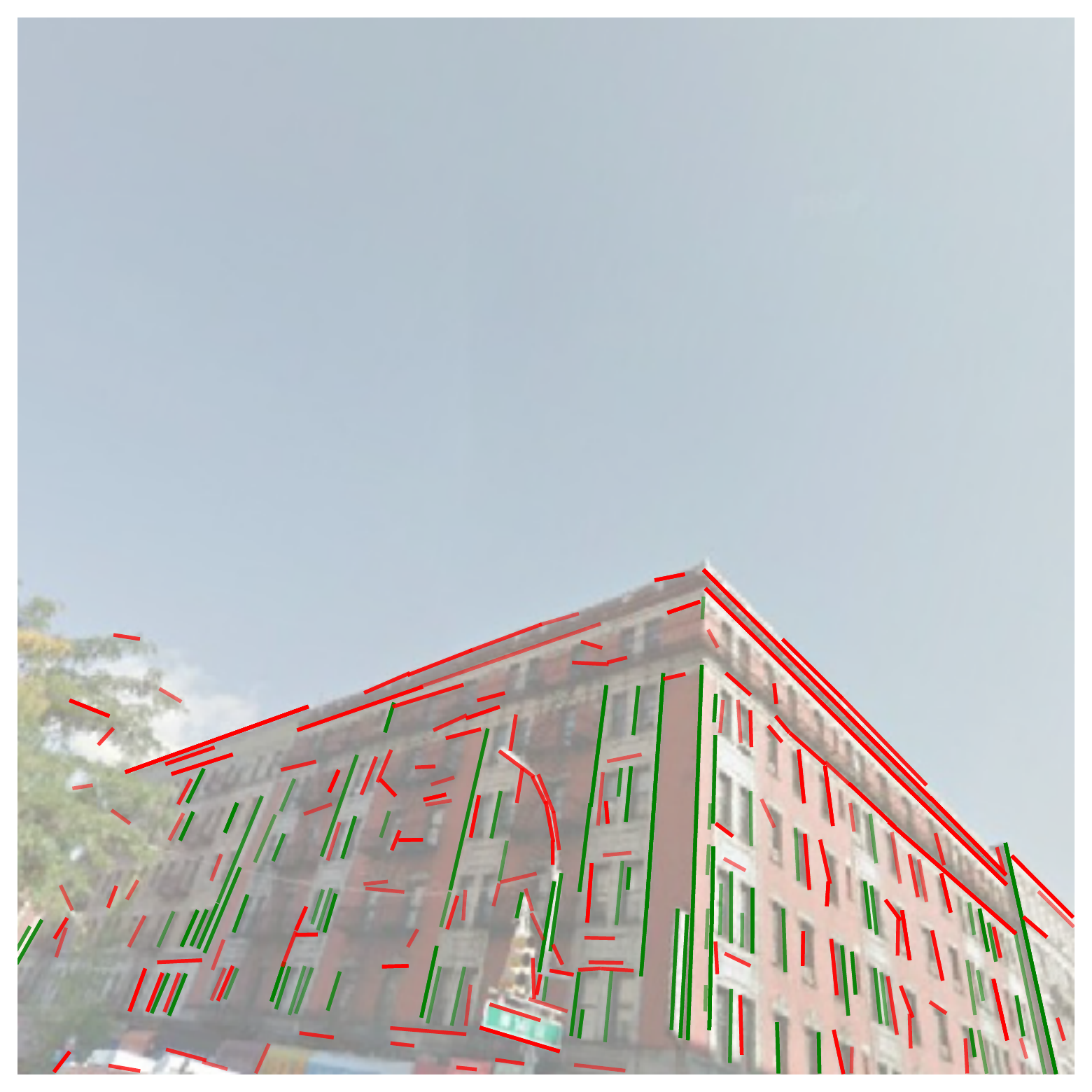} &
    \includegraphics[width=0.24\linewidth]{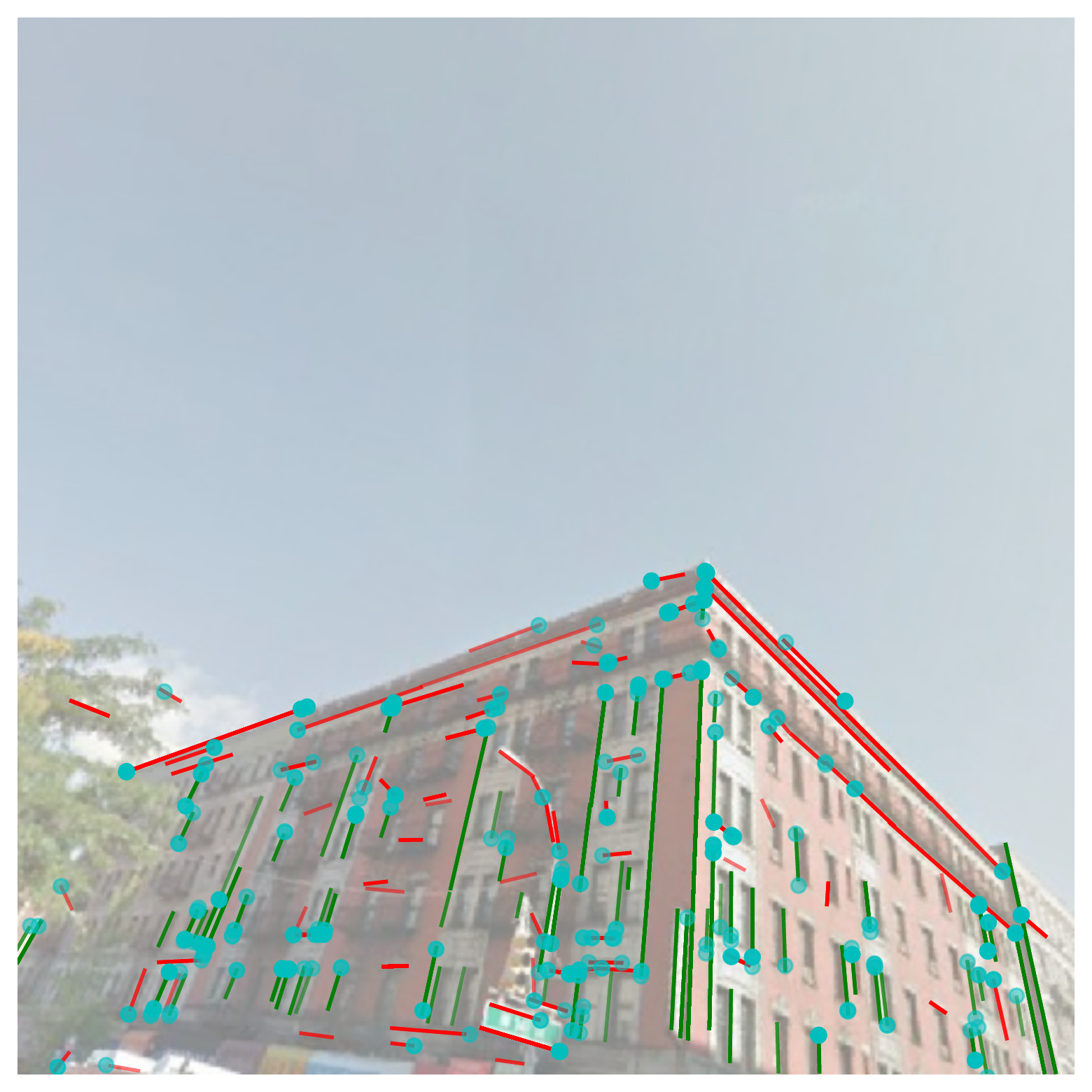} &
    \includegraphics[width=0.24\linewidth]{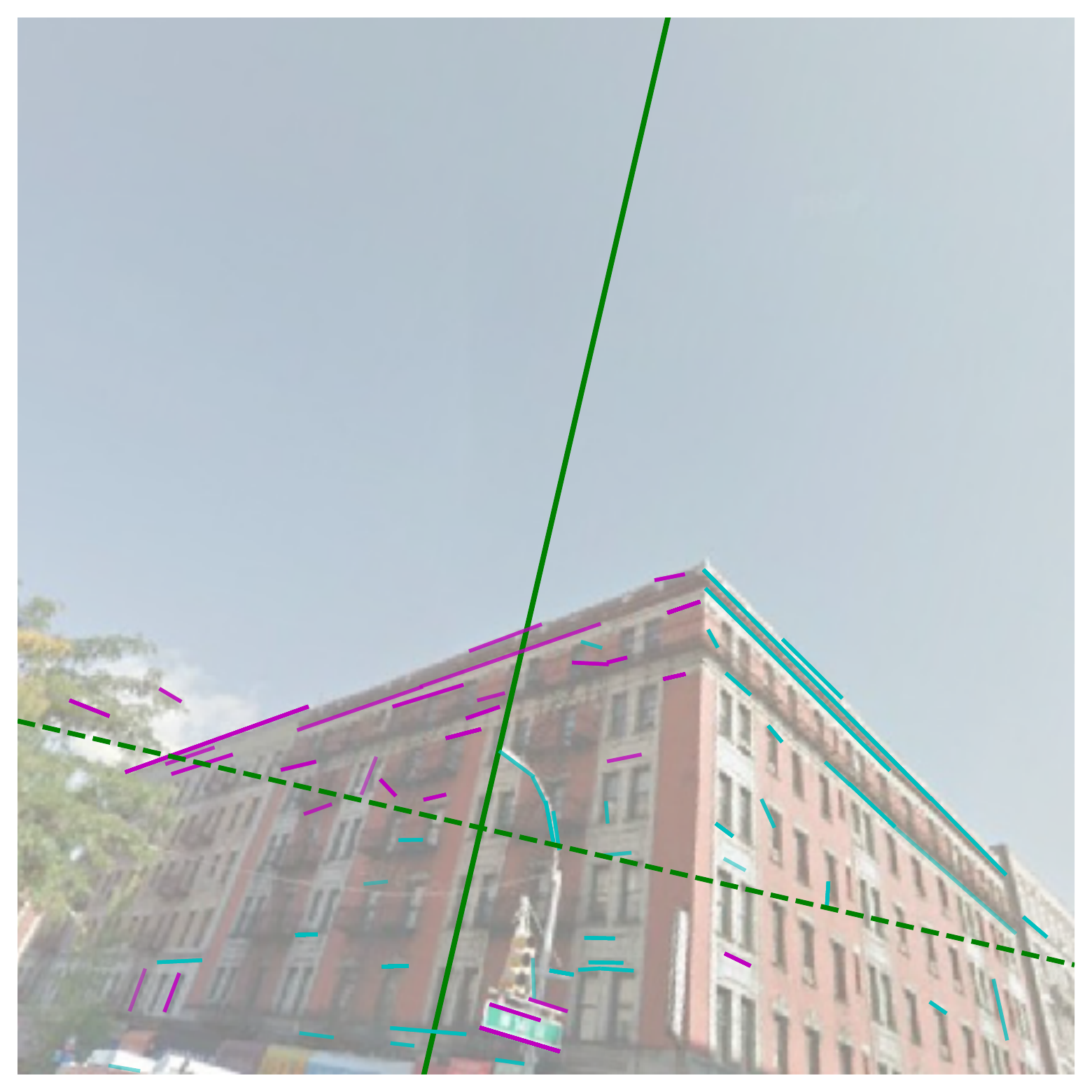} \\
    (a) & (b) & (c) & (d)
    \end{tabular}
    \caption{Sampling horizontal line segments into two groups:
    (a) Based on a zenith VP representative (green), we want to classify the line segments (blue) of a given image. 
    (b) Line segments are classified as follows: vanishing lines of the zenith VP (green) and the remaining lines (red).
    (c) Junction points (cyan) are computed as the intersections of spatially adjacent line segments that are classified differently; line segments whose endpoints are close to junction points are selected.
    (d) Using a pseudo-horizon (dotted line), we divide horizontal line segments into two groups (magenta and cyan).
    }
    \label{fig:junc_filter}
\end{figure}

After we extract a set of zenith candidates, we estimate the remaining two horizontal VPs taking into account the given set of zenith VP candidates. We first generate a set of hypotheses on all three VPs. Each hypothesis is then scored by our FSNet.

To sample horizontal VPs, we first filter the input line segments. However, we cannot simply filter line segments using their directions in this case, as there may be multiple horizontal VPs, and lines in any directions may vanish in the horizon. As a workaround, we use a heuristic based on the characteristics of most urban scenes.

Many man-made structures contain a large number of rectangles (e.g., facades or windows of a building) that are useful for calibration parameter estimation, and line segments enclosing these rectangles create junction points. Therefore, we sample horizontal direction line segments by only using their endpoints when they are close to the endpoints of the estimated vertical vanishing lines.

\Fig{junc_filter} illustrates the process of sampling horizontal line segments into two groups. Let $\mathbf{z}_{est} = (z_x, z_y, z_w)$ be a representative of the estimated zenith VPs, which is computed as the eigenvector with the largest eigenvalue of $\overline{\mathrm{ST}}(\mathbf{z})$ in \Eq{zenith_loc_loss}.
We first draw a pseudo-horizon by using $\mathbf{z}_{est}$ and then compute the intersection points between each sampled line segment and the pseudo-horizon. Finally, using a line connecting $\mathbf{z}_{est}$ and the image center as a pivot, we divide horizontal line segments into two groups; one that intersects the pseudo-horizon on the left side of the pivot and the other that intersects the pseudo-horizon on the right side of the pivot. The set of horizontal VP candidates is composed of intersection points by randomly sampling pairs of horizontal direction line segments in each group. We sample an equal number of candidates for both groups.

Once the set of horizontal VP candidates is sampled, we sample candidates of Manhattan directions. To sample each candidate, we draw two VPs; one from zenith candidates and the other from either set of horizontal VP candidates. The calibration parameters for the candidate can then be estimated by solving \Eq{camera_projection} with the two VPs, assuming that the principal point is on the image center \cite{Kosecka:2002}.

We design our FSNet for inferring camera calibration parameters to utilize all the available data, including VPs, lines, and the original raw image (\Fig{overview}, top-right). ResNet~\cite{He:2016} is adapted to our FSNet to handle raw images, appending all the other data as additional color channels. 
To append the information of the detected line segments, we rasterize line segments as a binary line segment map whose width and height are the same as those of the input image, as follows:
%
\begin{equation}
    \mathbf{L}(u, v) = \left\{ 
        \begin{array}{lll}
            1 & ~ & \mathrm{if~a~line~} \mathbf{l} \mathrm{~passes~through~} (u, v) \\
            0 & ~ & \mathrm{otherwise}
        \end{array}
    \right.,
    \label{eq:line_map}
\end{equation}
where $(u, v)$ represents a pixel location of the line segment map.
We also append the information of vanishing line segments (i.e., expanding lines are close to a VP) as a weighted line segment map for all three VPs of a candidate, where weights are computed using the closeness between the line segments and VPs. For a given VP $\mathbf{v}$ and the line equation $\mathbf{l}$ of a line segment, we compute the closeness between $\mathbf{v}$ and $\mathbf{l}$ using the conventional line-point distance as follows:
\begin{equation}
    \mathrm{closeness}\left( \mathbf{l}, \mathbf{v} \right) = 1 - \frac{\left| \mathbf{l} \cdot \mathbf{v} \right|}{\left\Vert \mathbf{l} \right\Vert \left\Vert \mathbf{v} \right\Vert}.
\label{eq:line_segment_map}
\end{equation}
Three activation maps are drawn for each candidate ($x$-, $y$- and $z$-directions), as:
\begin{equation}
    \mathbf{A}_{\left\{ x|y|z \right\}}(u, v) = \left\{ 
        \begin{array}{lll}
            \mathrm{closeness}(\mathbf{l}, \mathbf{v}_{\left\{ x|y|z \right\}}) &  & \mathrm{if~a~line~} \mathbf{l} \mathrm{~passes~through~} (u, v) \\
            0 &  & \mathrm{otherwise}
        \end{array}
    \right..
\label{eq:closeness_map}
\end{equation}

All the maps are appended to the original image as illustrated in \Fig{overview}. Finally, we append the Manhattan directions and the estimated focal length to each pixel of the concatenated map so that the input to the scoring network have size of $(\emph{height}, \emph{width}, \emph{channel}) = (224, 224, 17)$ (\Fig{overview}, top-right).

To train FSNet, we assign GT score to each candidate by measuring similarities between horizon and zenith of each candidate and those of the GT. For the zenith, we measure the cosine similarities of GT zenith and that of candidate as follows:
\begin{equation}
    s_{z_i} = \mathrm{cossim}(\mathbf{z}_{gt}, \mathbf{z}_i),
\end{equation}
where $\mathbf{z}_{gt}$ and $\mathbf{z}_i$ represent the GT and candidate zenith. For the horizon, we adapt the distance metric proposed in \cite{Barinova:2010}. For this, we compute the intersection points between the GT and candidate horizons and left/right image boundaries. Let $\mathbf{h}_l$ and $\mathbf{h}_r$ be intersection points of the predicted horizon and left/right border of the image. Similarly, we compute $\mathbf{g}_l$ and $\mathbf{g}_r$ using the ground truth horizon. Inspired by \cite{Barinova:2010}, the similarity between the GT and a candidate is computed as:
\begin{equation}
s_{h_i} = \exp \left( -\max\left( \Vert \mathbf{h}_{l_i} - \mathbf{g}_l \Vert_1,
                             \Vert \mathbf{h}_{r_i} - \mathbf{g}_r \Vert_1
                       \right)^2 \right).
\label{eq:similarity}
\end{equation}
\noindent
Our scoring network $h_{score}$ is then trained with the cross entropy loss, defined as:
\begin{align}
    \mathcal{L}_{score} &= \sum_i -h_{score} (\mathbf{R}_i) \log(c_i) \\
    c_i &= \left\{ 
        \begin{array}{lll}
            0 & ~~ & \mathrm{if~} s_{vh_i} < \delta_s \\
            1 & ~~ & \mathrm{otherwise}
        \end{array} \right.
        \label{eq:score_tol}\\
    s_{vh_i} &= \exp \left( -\frac{\left( \frac{s_{h_i} + s_{v_i}}{2} - 1.0 \right)^2}{2\sigma^2} \right),
\end{align}
where $\sigma = 0.1$ and $\delta_s = 0.5$ in our experiments.

\subsubsection{Robust score estimation using the Manhattan world assumption.}
Although our FSNet is able to accurately estimate camera calibration parameters in general, it can sometimes be noisy and unstable. In our experiments, we found that incorporating with the Manhattan world assumption increased the robustness of our network. Given a line segment map and three closeness maps (\Eqs{line_segment_map} and (\ref{eq:closeness_map})), 
we compute the extent to which a candidate follows the Manhattan world assumption using \Eq{activation_map_score}:
\begin{equation}
    m_i = \frac{\sum_u \sum_v 
                \max \left( \mathbf{A}_x(u, v), \mathbf{A}_y(u, v), \mathbf{A}_z(u, v) \right)}
               {\sum_u \sum_v \mathbf{L}(u, v)},
    \label{eq:activation_map_score}
\end{equation}
and the final score of a candidate is computed as:
\begin{equation}
    s_i = s_{vh_i} \cdot m_i.
    \label{eq:final_score}
\end{equation}
\noindent
Once all the candidates are scored, we estimate the final focal length and zenith by averaging those of top-$k$ high score candidates such that:
\begin{equation}
    f_{est} = \frac{\sum_i s_i f_i}{\sum_i s_i} \mathrm{~~and~~}
    \overline{\mathrm{ST}}_{est} = \frac{\sum_i s_i \mathrm{ST}(z_i)}{\sum_i s_i},
    \label{eq:avg_top_k}
\end{equation}
where $\mathrm{ST}(\mathbf{z}_i)$ represents the structure tensor of a zenith candidate (\Eq{structure_tensor}). We set $k=8$ in the experiments.
$\mathbf{z}_{est}$ can be estimated from $\overline{\mathrm{ST}}_{est}$,
and the two camera rotation angles $\psi$ and $\phi$ in \Eq{camera_projection} can be computed from $f_{est}$ and $\mathbf{z}_{est}$. For the rotation angle $\theta$, we simply take the value from the highest score candidate, as there may be multiple pairs of horizontal VPs that are not close to each other, particularly when the scene does not follow the Manhattan world assumption but the Atlanta world assumption. Note that they still share similar zeniths and focal lengths \cite{Barinova:2010,Lee:2014}.

\subsection{Training Details and Runtime}

In training, we train ZSNet first and then train FSNet with the outputs of ZSNet. Both networks are trained with Adam optimizer with initial learning rates of 0.001 and 0.0004 for ZSNet and FSNet, respectively. Both learning rates are decreased by half for every 5 epochs. The mini-batch sizes of ZSNet and FSNet are set to 16 and 2, respectively. The input images are always downsampled to $224 \times 224$, and the LSD \cite{Gioi:2010} is computed on these low-res images. 

At test time, it takes $\sim$0.08s for data preparation with Intel Core i7-7700K CPU and another $\sim$0.08s for ZSNet/FSNet per image with Nvidia GeForce GTX 1080 Ti GPU.

%% file: sections/experiments.tex

\section{Experiments}
\label{sec:experiments}
We provide the experimental results with Google Street View~\cite{GSVI} and HLW~\cite{Workman:2016} datasets. Refer to the supplementary material for more experimental results with the other datasets and more qualitative results.

\parahead{Google Street View~\cite{GSVI} dataset} It provides panoramic images of outdoor city scenes for which the Manhanttan assumption is satisfied. For generating training and test data, we first divide the scenes for each set and rectify and crop randomly selected panoramic images by sampling FoV, pitch, and roll in the ranges of $40 \sim 80 ^{\circ}$, $-30 \sim 40^{\circ}$, and $-20 \sim 20^{\circ}$, respectively. 13,214 and 1,333 images are generated for training and test sets, respectively.

\parahead{HLW~\cite{Workman:2016} dataset} 
It includes images only with the information of horizon line but no other camera intrinsic parameters. Hence, we use this dataset only for verifying the generalization capability of methods at test time.

\parahead{Evaluation Metrics}
We measure the accuracy of the output camera up vector, focal length, and horizon line with several evaluation metrics. For camera up vector, we measure the difference of angle, pitch, and roll with the GT. For the focal length, we first convert the output focal length to FoV and measure the angle difference with the GT. Lastly, for the horizon line, analogous to our similarity definition in \Eq{similarity}, we measure the distances between the predicted and GT lines at the left/right boundary of the input image (normalized by the image height) and take the maximum of the two distances. We also report the area under curve (AUC) of the cumulative distribution with the $x$-axis of the distance and the $y$-axis of the percentage, as introduced in \cite{Barinova:2010}. The range of $x$-axis is $[0, 0.25]$.

\subsection{Comparisons}
\label{sec:comparisons}

\begin{table}[t!]
\renewcommand{\arraystretch}{0.9}
\caption{Supervision and output characteristics of baseline methods and ours. The first two are unsupervised methods not leveraging neural networks, and the others are deep learning methods. Ours is the only network-based method predicting all four outputs.}
\label{tbl:supervision}
{\scriptsize
\begin{tabularx}{\textwidth}{l|CCCC|CCCC}
\toprule
Method & \multicolumn{4}{c|}{Supervision} & \multicolumn{4}{c}{Output} \\
\cline{2-9} & Horizon Line & Focal Length & Camera Rotation & Per-Pixel Normal & Horizon Line & Focal Length & Camera Rotation & Up Vector \\ 
\midrule
Upright~\cite{Lee:2014}                       & \multicolumn{4}{c|}{\multirow{2}{*}{N/A}} & \checkmark & \checkmark & \checkmark & \checkmark \\
A-Contrario~\cite{Simon:2018}                 & \multicolumn{4}{c|}{} & \checkmark & \checkmark & \checkmark & \checkmark \\
\midrule
DeepHorizon~\cite{Workman:2016}               & \checkmark &            &            &            & \checkmark &            &            &            \\
Perceptual~\cite{Hold-Geoffroy:2018}          & \checkmark & \checkmark &            &            & \checkmark & \checkmark &            & \checkmark \\
UprightNet~\cite{Xian:2019}                   &            &            & \checkmark & \checkmark &            &            &            & \checkmark \\
\midrule
\textbf{Ours}                                 & \checkmark & \checkmark &            &            & \checkmark & \checkmark & \checkmark & \checkmark \\
\bottomrule
\end{tabularx}
}
\end{table}

We compare our method with six baseline methods, where \Tbl{supervision} presents the required supervision characteristics and outputs. Upright~\cite{Lee:2014} and A-Contrario Detection~\cite{Simon:2018} are non-neural-net methods based on line detection and RANSAC. We use the authors' implementations in our experiments. For Upright \cite{Lee:2014}, the evaluation metrics are applied after optimizing the Manhattan direction per image, assuming the principal point as the image center. A-Contrario Detection~\cite{Simon:2018} often fails to estimate focal length when horizon VP candidates are insufficient in the input. Thus, we exclude \cite{Simon:2018} in the evaluation of FoV. The AUC of \cite{Simon:2018} is measured, regardless of the failures in focal length estimations, with the horizon lines estimated by the method proposed in \cite{Simon:2018}. The other metrics of \cite{Simon:2018} are evaluated only for the cases that focal lengths are successfully estimated. DeepHorizon~\cite{Workman:2016} and Perceptual Measure~\cite{Hold-Geoffroy:2018} are neural-network-based methods directly taking the global feature of an image and performing classifications in discretized camera parameter spaces. For fair comparisons, we use ResNet~\cite{He:2016} as a backbone architecture in the implementations of these methods. Note that DeepHorizon~\cite{Workman:2016} does not predict focal length, and thus we use ground truth focal length in the estimation of the camera up vector. Additionally, we train Perceptual Measure~\cite{Hold-Geoffroy:2018} by feeding it both the input image and the line map used in our FSNet (\Sec{fsnet}), and we assess whether the extra input improves the performance. UprightNet~\cite{Xian:2019} is another deep learning method that requires additional supervision in training, such as camera extrinsic parameters and per-pixel normals in the 3D space. Due to the lack of such supervision in our datasets, in our experiments, we use the author's pretrained model on ScanNet~\cite{ScanNet:2017}, which is a synthetic dataset.

\begin{table}[t!]
\renewcommand{\arraystretch}{0.9}
\caption{Quantitative evaluations with Google Street View dataset~\cite{GSVI}. See \textbf{Evaluation Metrics} in \Sec{experiments} for details. Bold is the best result, and underscore is the second-best result. Note that, for DeepHorizon~\cite{Workman:2016}*, we use GT FoV to calculate the camera up vector (angle, pitch, and roll errors) from the predicted horizon line. Also, for UprightNet~\cite{Xian:2019}**, we use a pretrained model on ScanNet~\cite{ScanNet:2017} due to the lack of required supervision in the Google Street View dataset.}
\label{tbl:test_googlestreetview}
{\scriptsize
\begin{tabularx}{\textwidth}{l|CC|CC|CC|CC|C}
\toprule
\multirow{2}{*}{Method} & \multicolumn{2}{c|}{Angle ($^\circ$) $\downarrow$}  & \multicolumn{2}{c|}{Pitch ($^\circ$) $\downarrow$} & \multicolumn{2}{c|}{Roll ($^\circ$) $\downarrow$} & \multicolumn{2}{c|}{FoV ($^\circ$) $\downarrow$}  & \multirow{2}{*}{\makecell{AUC\\($\%$) $\uparrow$}} \\
\cline{2-9} & Mean & Med. & Mean & Med. & Mean & Med. & Mean & Med. \\ 
\midrule
Upright~\cite{Lee:2014}                             &  3.05 &  1.92 &  2.90 &  1.80 &  6.19 & \textbf{0.43} & 9.47 &  4.42 & 77.43 \\
A-Contrario~\cite{Simon:2018}                       &  3.93 & \underline{1.85} &  3.51 &  \underline{1.64} & 13.98 &  0.52 &   -   &   -    & 74.25 \\
DeepHorizon~\cite{Workman:2016}*                    &  3.58 &  3.01 &  2.76 &  2.12 &  1.78 &  1.67 &   -   &   -    & 80.29 \\
Perceptual~\cite{Hold-Geoffroy:2018}                &  2.73 &  2.13 &  2.39 &  1.78 &  0.96 &  0.66 & \textbf{4.61} &  \underline{3.89}  & \underline{80.40} \\
Perceptual~\cite{Hold-Geoffroy:2018} $+ \mathbf{L}$ & \underline{2.66} &  2.10 &  \underline{2.31} &  1.80 &  \underline{0.92} &  0.93 & \underline{5.27} &  3.99  & \underline{80.40} \\
UprightNet~\cite{Xian:2019}**                       & 28.20 & 26.10 & 26.56 & 24.56 &  6.22 &  4.33 &   -   &   -   &   -   \\
\midrule
\textbf{Ours}                               & \textbf{2.12} & \textbf{1.61} & \textbf{1.92} & \textbf{1.38} & \textbf{0.75} & \underline{0.47} &  6.01 & \textbf{3.72} & \textbf{83.12} \\
\bottomrule
\end{tabularx}
}
\end{table}

The quantitative results with Google Street View dataset~\cite{GSVI} are presented in~\Tbl{test_googlestreetview}. The results demonstrate that our method outperforms all the baseline methods in most evaluation metric. Upright~\cite{Lee:2014} provides a slightly lower median roll than ours, although its mean roll is much greater than the median, meaning that it completely fails in some test cases. In addition, Perceptual Measure~\cite{Hold-Geoffroy:2018} gives a slightly smaller mean FoV error; however, the median FoV error is higher than ours. When Perceptual Measure~\cite{Hold-Geoffroy:2018} is trained with the additional line map input, the result indicates that it does not lead to a meaning difference in performance. As mentioned earlier, a pretrained model is used for UprightNet~\cite{Xian:2019} (trained on ScanNet~\cite{ScanNet:2017}) due to the lack of required supervision in Google Street View; thus the results are much poorer than others. 

\begin{figure*}[t!]
    \centering
    \begin{tabular}{ccccc}
    \includegraphics[width=0.18\linewidth]{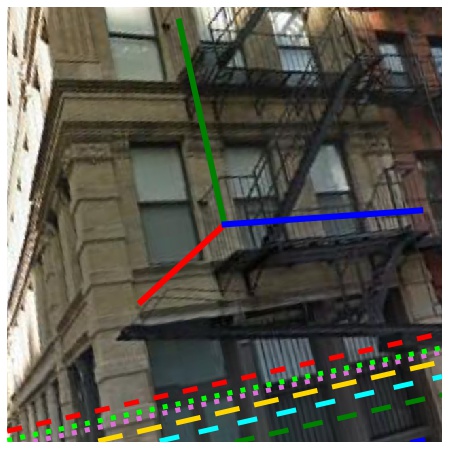} &
    \includegraphics[width=0.18\linewidth]{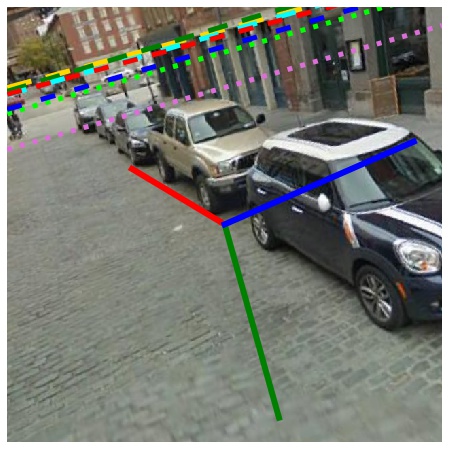} &
    \includegraphics[width=0.18\linewidth]{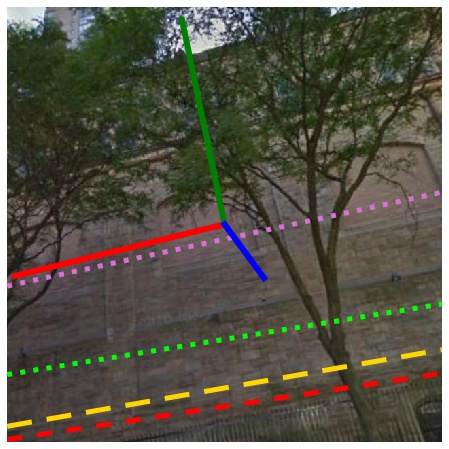} &
    \includegraphics[width=0.18\linewidth]{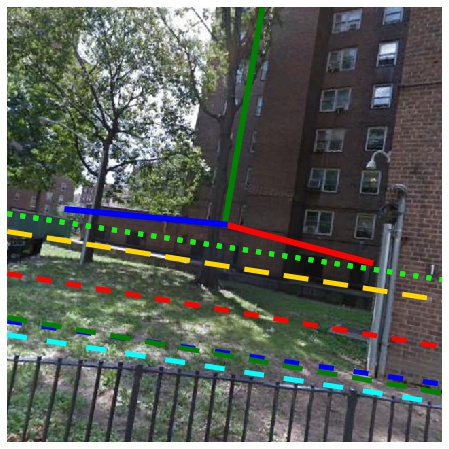} &
    \includegraphics[width=0.18\linewidth]{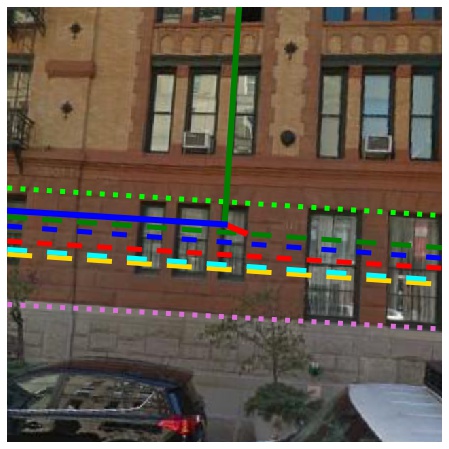} \\
    \includegraphics[width=0.18\linewidth]{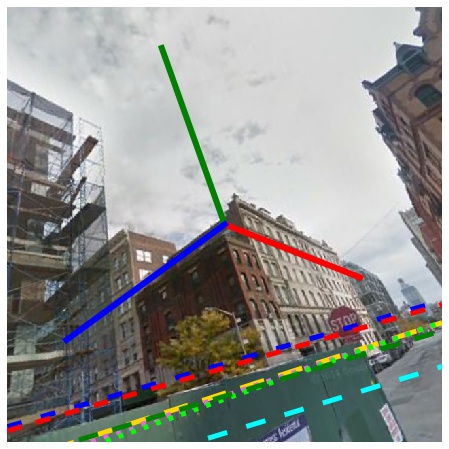} &
    \includegraphics[width=0.18\linewidth]{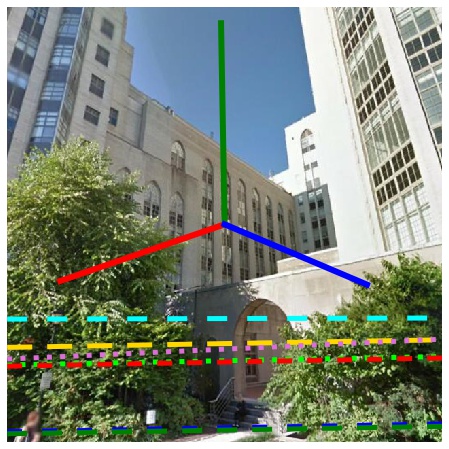} &
    \includegraphics[width=0.18\linewidth]{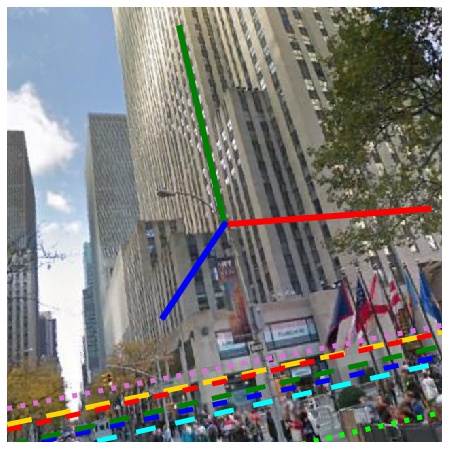} &
    \includegraphics[width=0.18\linewidth]{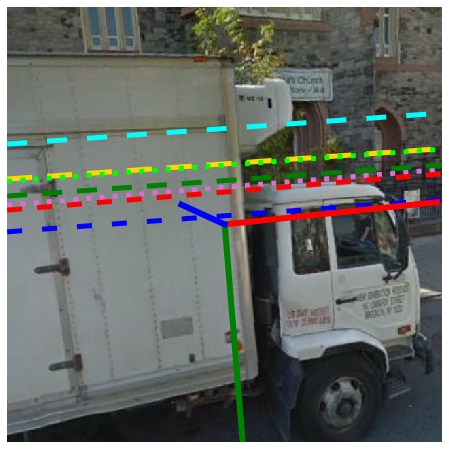} &
    \includegraphics[width=0.18\linewidth]{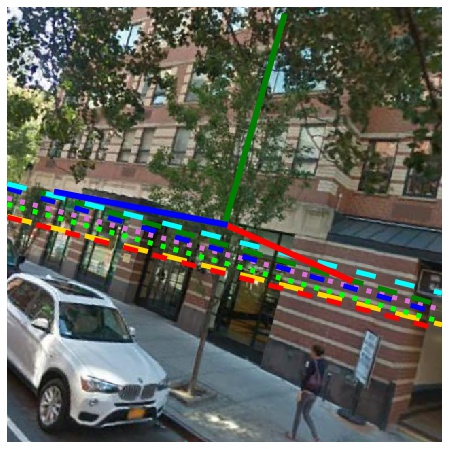} \\
    \includegraphics[width=0.18\linewidth]{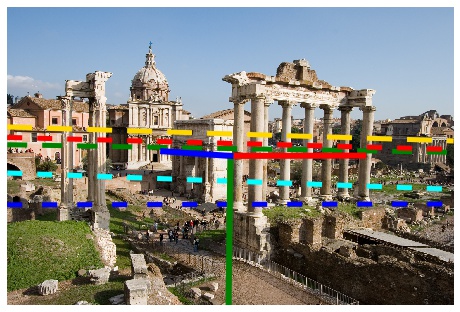} &
    \includegraphics[width=0.18\linewidth]{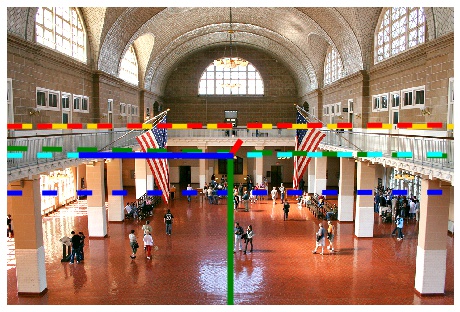} &
    \includegraphics[width=0.18\linewidth]{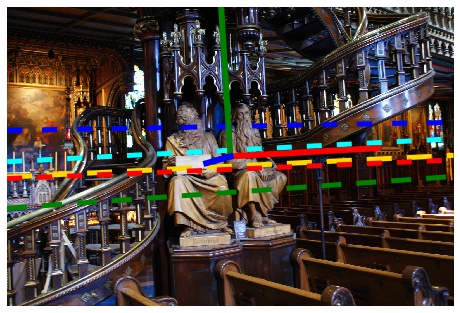} &
    \includegraphics[width=0.18\linewidth]{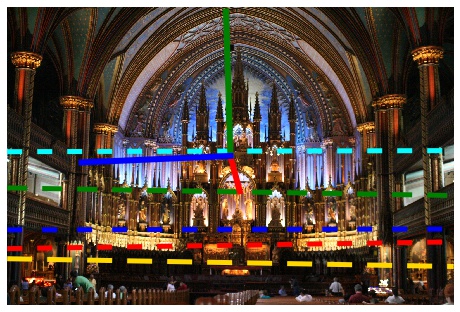} &
    \includegraphics[width=0.18\linewidth]{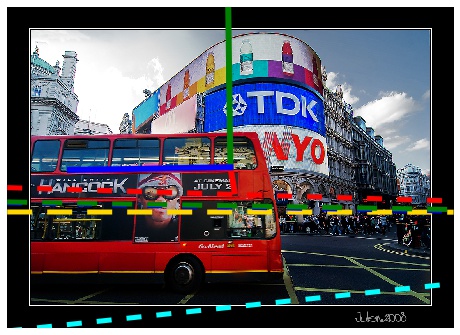} \\
    \includegraphics[width=0.18\linewidth]{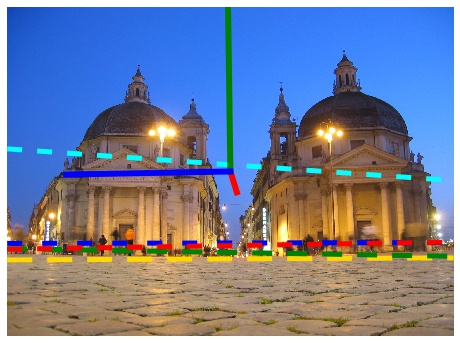} &
    \includegraphics[width=0.18\linewidth]{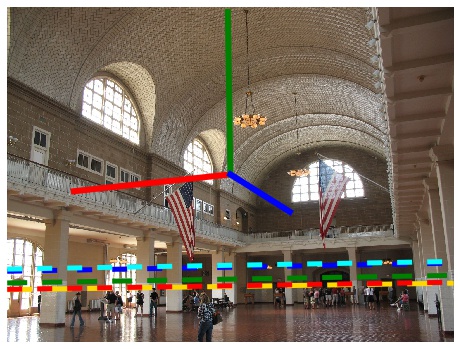} &
    \includegraphics[width=0.18\linewidth]{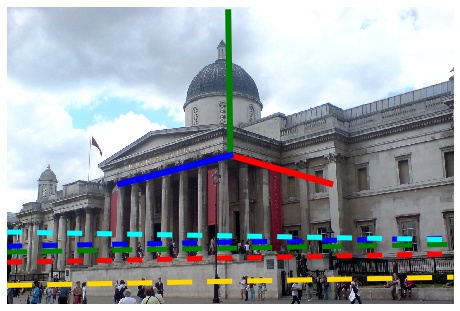} &
    \includegraphics[width=0.18\linewidth]{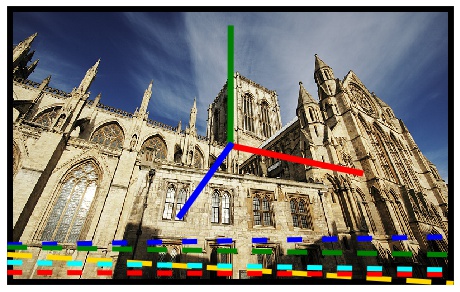} &
    \includegraphics[width=0.18\linewidth]{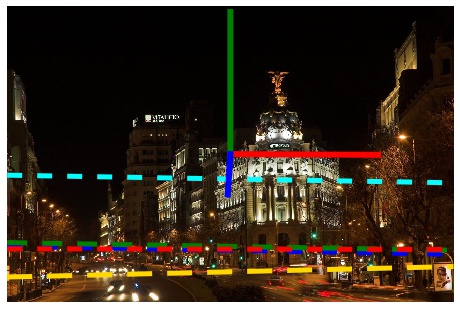} \\
    \end{tabular}
    \newcommand{\crule}[3][red]{\textcolor{#1}{\rule{#2}{#3} \rule{#2}{#3} \rule{#2}{#3} \rule{#2}{#3}}}
    {\scriptsize
    \begin{tabular}{llll}
    \crule[Goldenrod]{0.01\linewidth}{0.01\linewidth} Ground Truth & 
    \crule[Orchid]{0.01\linewidth}{0.01\linewidth} Upright \cite{Lee:2014} & 
    \crule[LimeGreen]{0.01\linewidth}{0.01\linewidth} A-Contario \cite{Simon:2018} & 
    \crule[blue]{0.01\linewidth}{0.01\linewidth} DeepHorizon \cite{Workman:2016} \\ 
    &
    \crule[OliveGreen]{0.01\linewidth}{0.01\linewidth} Perceptual \cite{Hold-Geoffroy:2018} & 
    \crule[SkyBlue]{0.01\linewidth}{0.01\linewidth} Perceptual \cite{Hold-Geoffroy:2018} $+ \mathbf{L}$  &  
    \crule[red]{0.01\linewidth}{0.01\linewidth} Ours \\
    \end{tabular}
    }
    \caption{Examples of horizon line prediction on the Google Street View test set (top two rows) and on the HLW test set (bottom two rows). Each example also shows the Manhattan direction of the highest score candidate.}
    \label{fig:horizon_line_predictions}
\end{figure*}

\Fig{horizon_line_predictions} visualizes several examples of horizon line predictions as well as our weakly-supervised Manhattan directions. Recall that we do not use \emph{full} supervision of the Manhattan directions in training; we only use the supervision of horizon lines and focal lengths. In each example in \Fig{horizon_line_predictions}, we illustrate the Manhattan direction of the highest score candidate.

\begin{figure}[t!]
\setlength{\tabcolsep}{1em}
\centering
\begin{tabular}{cc}
    \includegraphics[width=0.4\textwidth]{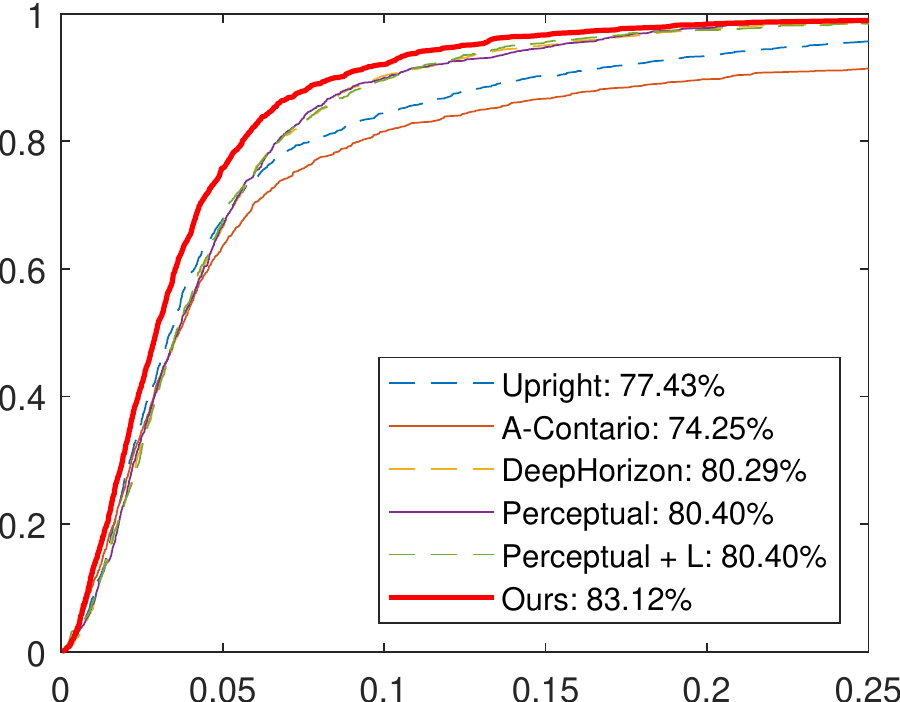} & 
    \includegraphics[width=0.4\textwidth]{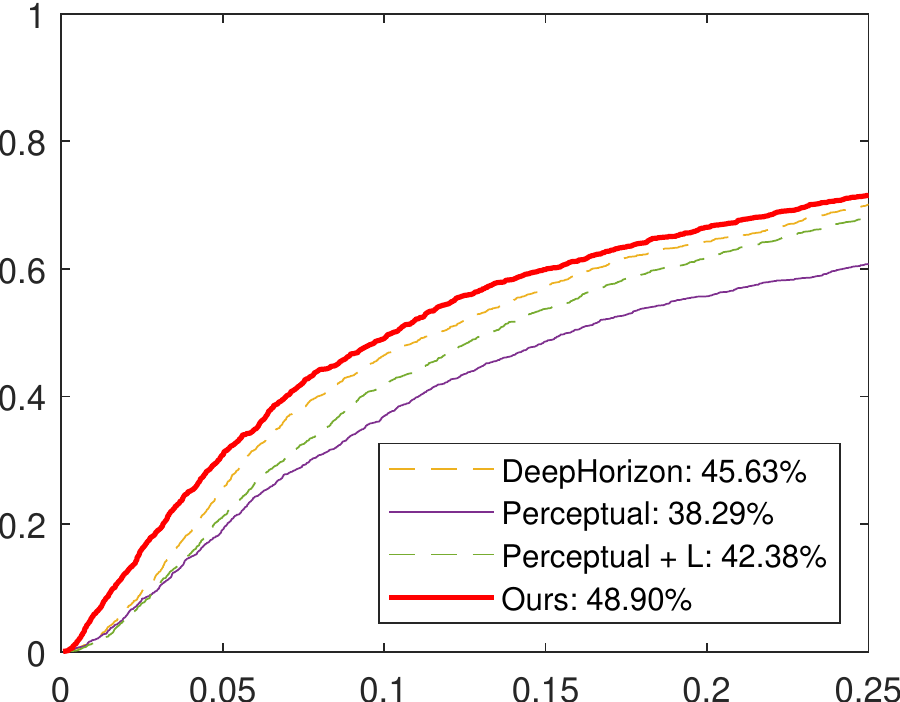} \\
    (a) Google Street View & 
    (b) HLW  \\
\end{tabular}
\caption{Comparison of the cumulative distributions of the horizon line error and their AUCs tested on (a) Google Street View and (b) HLW. Note that, in (b), neural approaches are trained with the Google Street View training dataset and to demonstrate the generalization capability. The AUCs in (a) are also reported in~\Tbl{test_googlestreetview}.}
\label{fig:auc}
\end{figure}

To evaluate the generalization capability of neural-network-based methods, we also take the network models trained on Google Street View training dataset and test them on the HLW dataset~\cite{Workman:2016}. Because the HLW dataset only has the GT horizon line, \Fig{auc}(b) only reports the cumulative distributions of the horizon prediction errors. As shown in the figure, our method provides the largest AUC with a significant margin compared with the other baselines. Interestingly, Perceptual Measure~\cite{Hold-Geoffroy:2018} shows improvement when trained with the additional line map, meaning that the geometric interpretation helps more when parsing \emph{unseen} images in network training.

\begin{figure}[t]
\begin{minipage}[c]{0.35\linewidth}
\renewcommand{\arraystretch}{0.9}
\captionof{table}{Evaluation of ZSNet.}
{\scriptsize
\begin{tabularx}{\linewidth}{l|CC}
\toprule
\multirow{2}{*}{} & \multicolumn{2}{c}{Angle ($^\circ$) $\downarrow$}   \\
\cline{2-3} & Mean & Med. \\ 
\midrule
ZSNet {\tiny (Ours)}          & \textbf{2.53} & \textbf{1.52} \\
Upright~\cite{Lee:2014}       & 3.15 & \underline{2.11} \\
A-Contrario~\cite{Simon:2018} & \underline{3.06} & 2.38 \\
\bottomrule
\end{tabularx}
\label{tbl:comp_zenith}
}
\end{minipage}
\hfill 
\begin{minipage}[c]{0.63\linewidth}
\begin{tabular}{cc|cc}
\includegraphics[width=0.23\linewidth]{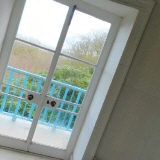} &
\includegraphics[width=0.23\linewidth]{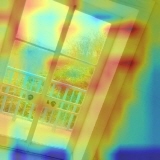} &
\includegraphics[width=0.23\linewidth]{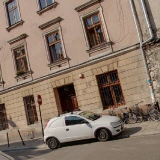} &
\includegraphics[width=0.23\linewidth]{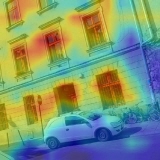}
\end{tabular}
\caption{Visualizations of FSNet focus: (left) input; (right) feature highlight.}
\label{fig:feature_highlight}
\end{minipage}
\end{figure}

We conduct an experiment comparing the outputs of ZSNet with the outputs of Upright \cite{Lee:2014} and A-Contrario \cite{Simon:2018}. Using the weighted average of the zenith candidates (in \Eq{zenith_loc_loss}), we measured the angle to the GT zenith $\mathrm{cossim}(z_i, z_{gt})$, as provided in \Eq{zenith_label}. \Tbl{comp_zenith} show that our ZSNet computes the zenith VP more accurately than the other non-neural-net methods.

\Fig{feature_highlight} visualizes the weights of the second last convolution layer in FSNet (the layer in the ResNet backbone); red means high, and blue means low. It can be seen that our FSNet focused on the areas with many line segments, such as buildings, window frames, and pillars. The supplementary material contains more examples.

\subsection{Ablation Study}
\label{sec:ablation_study}
We conduct an ablation study using Google Street View dataset to demonstrate the effect of each component in our framework. All results are reported in \Tbl{ablation}, where the last row shows the result of our final version framework.

We first evaluate the effect of the entire ZSNet by ablating it in the training. When sampling zenith candidates in FSNet (\Sec{fsnet}), the score $p_{z_i}$ in \Eq{zenith_pred} is not predicted but set uniformly. The first row of \Tbl{ablation} indicates that the performance significantly decreases in all evaluation metrics; e.g., the AUC decreased by 9\%. This indicates that ZSNet plays an important role in finding inlier Zenith VPs that satisfy the Manhattan/Atlanta assumption. In ZSNet, we also evaluate the effect of global line feature $\mathbf{g}_l$ in \Eq{line_feat} by not concatenating it with the point feature $h_z(\mathbf{z}_i)$ in \Eq{feat_concat}. 
Without the line feature, ZSNet is still able to prune the outlier zeniths in some extent, as indicated in the second row, but the performance is far inferior to that of our final framework (the last row). This result indicates that the equation of a horizon line is much more informative than the noisy coordinates of the zenith VP.

In FSNet, we first ablate some parts of the input fed to the network per frame. When we do not provide the given image but the rest of the input (the third row of \Tbl{ablation}), the performance decreases somewhat; however, the change is less significant than when omitting the line map $\mathbf{L}$ (\Eq{line_map}) and the activation map $\mathbf{A}$ (\Eq{closeness_map}) in the input (the fourth row). This demonstrates that FSNet learns more information from the line map and activation map, which contain explicit geometric interpretations of the input image. The combination of the two maps (our final version) produces the best performance. In addition, the results get worse when the activation map score $m_i$ is not used in the final score of candidates --- i.e., $s_i = s_{vh_i}$ in \Eq{final_score} (the fifth row).

\begin{table}[t!]
\renewcommand{\arraystretch}{0.9}
\caption{Ablation study results. Bold is the best result. See \Sec{ablation_study} for details.}
\label{tbl:ablation}
{\scriptsize
\begin{tabularx}{\textwidth}{c|l|CC|CC|CC|CC|C}
\toprule
\multicolumn{2}{l|}{} & \multicolumn{2}{c|}{Angle ($^\circ$) $\downarrow$}  & \multicolumn{2}{c|}{Pitch ($^\circ$) $\downarrow$} & \multicolumn{2}{c|}{Roll ($^\circ$) $\downarrow$} & \multicolumn{2}{c|}{FoV ($^\circ$) $\downarrow$}  & \multirow{2}{*}{\makecell{AUC\\($\%$) $\uparrow$}} \\
\cline{3-10}
\multicolumn{2}{c|}{} & Mean & Med. & Mean & Med. & Mean & Med. & Mean & Med. & \\ 
\midrule
\multicolumn{2}{c|}{w/o ZSNet}
    &  3.00 &  2.04 &  2.81 &  1.98 &  1.62 &  0.95 & 8.42 & 4.47 & 74.01 \\
\multicolumn{2}{c|}{$h'_z(\mathbf{z}_i) = h_z(\mathbf{z}_i)$ (\Eq{feat_concat})}
    &  4.34 &  1.96 &  3.91 &  1.76 &  1.64 &  0.59 & 7.88 & 4.16 & 77.65 \\
\midrule
\multicolumn{2}{c|}{FSNet$-$Image}
    &  2.45 &  1.78 &  2.19 &  1.52 & \textbf{0.68} & \textbf{0.47} &  6.71 &  4.35 & 80.20 \\
\multicolumn{2}{c|}{FSNet$-\mathbf{L}-\mathbf{A}$}
    &  3.74 &  2.22 &  3.09 &  1.91 &  1.68 &  0.66 & 8.26 & 5.40 & 74.31 \\
\multicolumn{2}{c|}{$s_i = s_{vh_i}$ (\Eq{final_score})}
    & 2.32 &  1.80 & 2.09 &  1.57 & 0.72 &  0.54 &  6.06 &  4.12 & 80.85 \\
\midrule
\multicolumn{2}{c|}{\textbf{Ours}}
    & \textbf{2.12} & \textbf{1.61} & \textbf{1.92} & \textbf{1.38} & 0.75 & \textbf{0.47} &  \textbf{6.01} & \textbf{3.72} & \textbf{83.12} \\
\bottomrule
\end{tabularx}
}
\end{table}

%% file: sections/conclusion.tex

\section{Conclusion}
\label{sec:conclusion}

In this paper, we introduced a neural method that predicts camera calibration parameters from a single image of a man-made scene. Our method fully exploits line segments as prior knowledge of man-made scenes, and in our experiments, it exhibited better performance than that of previous approaches. Furthermore, compared to previous neural approaches, our method demonstrated a higher generalization capability to unseen data. In future work, we plan to investigate neural camera calibration that considers a powerful but small number of geometric cues through analyzing image context, as humans do.

%% file: sections/appendix.tex

\ifpaper
  \newcommand\refpaper[1]{\unskip}
\else
  \makeatletter
  \newcommand{\manuallabel}[2]{\def\@currentlabel{#2}\label{#1}}
  \makeatother
  \manuallabel{sec:experiments}{4}
  \manuallabel{sec:comparisons}{4.1}
  \manuallabel{tbl:test_googlestreetview}{2}
  \manuallabel{fig:teaser}{1}
  \manuallabel{fig:auc}{4}
  \manuallabel{fig:horizon_line_predictions}{5}
  \manuallabel{fig:feature_highlight}{6}
  \manuallabel{eq:final_score}{22}
  \manuallabel{eq:angle_threshold}{4}
  \manuallabel{eq:zenith_label}{8}
  \manuallabel{eq:zenith_score}{12}
  \manuallabel{eq:score_tol}{19}
  \newcommand{\refpaper}[1]{in the paper}
\fi

\subsection{Comparisons on SUN360~\cite{SUN360:2012} Dataset}

Similar to the experiment in~\Sec{comparisons}~\refpaper{}, we also compare our method with the baseline methods using SUN360~\cite{SUN360:2012} dataset. We selected indoor and outdoor scenes in the SUN360 dataset that satisfied the Manhattan/Atlanta assumptions, and generated the training and test images in the same way as with Google Street View dataset, described in \Sec{experiments}~\refpaper{}. $30,837$ and $878$ images are generated for the training and test sets, respectively. Details of the evaluation metrics and baseline methods are provided in~\Sec{comparisons}~\refpaper{}.

\begin{figure*}[t!]
    \begin{center}
    \begin{tabular}{ccccc}
    \includegraphics[width=0.18\linewidth]{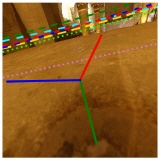} &
    \includegraphics[width=0.18\linewidth]{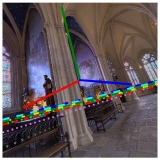} &
    \includegraphics[width=0.18\linewidth]{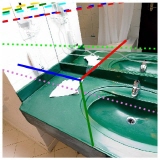} &
    \includegraphics[width=0.18\linewidth]{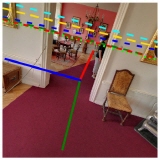} &
    \includegraphics[width=0.18\linewidth]{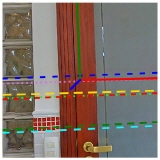} \\
    \includegraphics[width=0.18\linewidth]{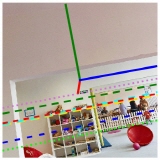} &
    \includegraphics[width=0.18\linewidth]{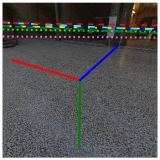} &
    \includegraphics[width=0.18\linewidth]{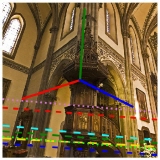} &
    \includegraphics[width=0.18\linewidth]{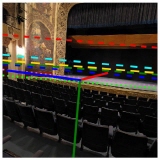} &
    \includegraphics[width=0.18\linewidth]{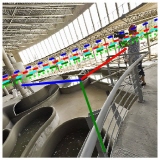} \\
    \includegraphics[width=0.18\linewidth]{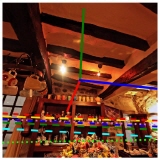} &
    \includegraphics[width=0.18\linewidth]{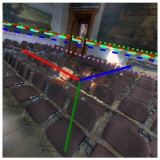} &
    \includegraphics[width=0.18\linewidth]{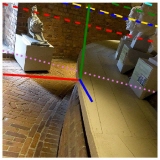} &
    \includegraphics[width=0.18\linewidth]{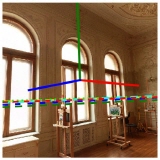} &
    \includegraphics[width=0.18\linewidth]{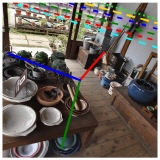} \\
    \includegraphics[width=0.18\linewidth]{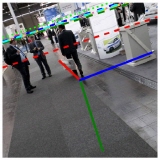} &
    \includegraphics[width=0.18\linewidth]{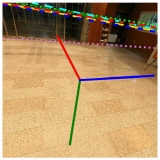} &
    \includegraphics[width=0.18\linewidth]{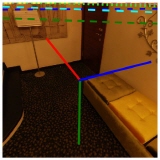} &
    \includegraphics[width=0.18\linewidth]{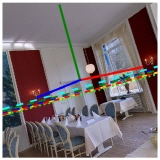} &
    \includegraphics[width=0.18\linewidth]{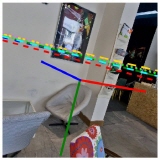} \\
    \includegraphics[width=0.18\linewidth]{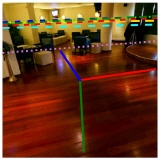} &
    \includegraphics[width=0.18\linewidth]{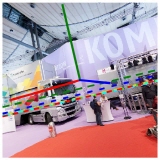} &
    \includegraphics[width=0.18\linewidth]{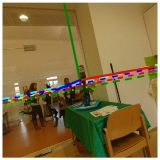} &
    \includegraphics[width=0.18\linewidth]{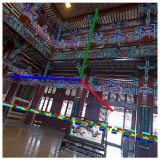} &
    \includegraphics[width=0.18\linewidth]{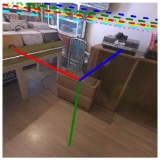} \\
    \includegraphics[width=0.18\linewidth]{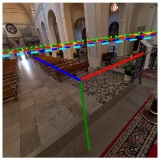} &
    \includegraphics[width=0.18\linewidth]{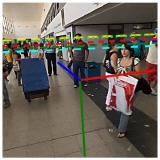} &
    \includegraphics[width=0.18\linewidth]{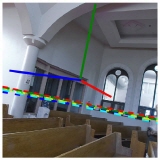} &
    \includegraphics[width=0.18\linewidth]{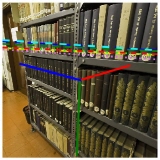} &
    \includegraphics[width=0.18\linewidth]{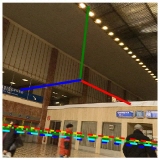} \\
    \includegraphics[width=0.18\linewidth]{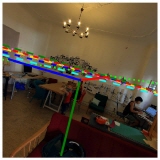} &
    \includegraphics[width=0.18\linewidth]{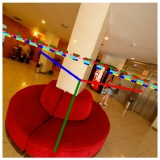} &
    \includegraphics[width=0.18\linewidth]{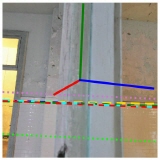} &
    \includegraphics[width=0.18\linewidth]{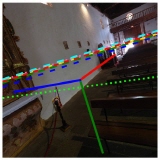} &
    \includegraphics[width=0.18\linewidth]{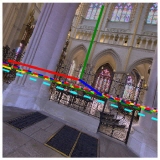} \\
    \end{tabular}
    \newcommand{\crule}[3][red]{\textcolor{#1}{\rule{#2}{#3} \rule{#2}{#3} \rule{#2}{#3} \rule{#2}{#3}}}
    {\scriptsize
    \begin{tabular}{llll}
    \crule[Goldenrod]{0.01\linewidth}{0.01\linewidth} Ground Truth & 
    \crule[Orchid]{0.01\linewidth}{0.01\linewidth} Upright \cite{Lee:2014} & 
    \crule[LimeGreen]{0.01\linewidth}{0.01\linewidth} A-Contario \cite{Simon:2018} & 
    \crule[blue]{0.01\linewidth}{0.01\linewidth} DeepHorizon \cite{Workman:2016} \\ 
    &
    \crule[OliveGreen]{0.01\linewidth}{0.01\linewidth} Perceptual \cite{Hold-Geoffroy:2018} & 
    \crule[SkyBlue]{0.01\linewidth}{0.01\linewidth} Perceptual \cite{Hold-Geoffroy:2018} $+ \mathbf{L}$  &  
    \crule[red]{0.01\linewidth}{0.01\linewidth} Ours \\
    \end{tabular}
    }
    \end{center}
    \caption{Examples of horizon line prediction on the SUN360 test set. Each example also displays the Manhattan direction of the highest score candidate.}
    \label{fig:horizon_line_predictions_sun360}
\end{figure*}

The quantitative results with SUN360 dataset~\cite{SUN360:2012} are reported in~\Tbl{test_sun360}. The trends of the results are similar to those of the Google Street View experiment in~\Tbl{test_googlestreetview}~\refpaper{}. Our method provides the best performance for most of the evaluation metrics, and the second-best for the remaining evaluation metrics, such as the median roll error and mean FoV error. Our method has a very marginal difference with the best AUC. The qualitative results are presented in~\Fig{horizon_line_predictions_sun360}.

\begin{table}[ht!]
\renewcommand{\arraystretch}{0.9}
\caption{Quantitative evaluations with SUN360 dataset. Bold represents the best result, while an underscore represents the second-best result. Note that for DeepHorizon~\cite{Workman:2016}*, we use the GT FoV to calculate the camera up vector (angle, pitch, and roll errors) from the predicted horizon line. In addition, for UprightNet~\cite{Xian:2019}**, we use a pretrained model on ScanNet~\cite{ScanNet:2017} due to the lack of required supervision in the SUN360 dataset.}
\vspace{-0.5\baselineskip}
\label{tbl:test_sun360}
{\scriptsize
\begin{tabularx}{\textwidth}{l|CC|CC|CC|CC|C}
\toprule
\multirow{2}{*}{Method} & \multicolumn{2}{c|}{Angle ($^\circ$) $\downarrow$}  & \multicolumn{2}{c|}{Pitch ($^\circ$) $\downarrow$} & \multicolumn{2}{c|}{Roll ($^\circ$) $\downarrow$} & \multicolumn{2}{c|}{FoV ($^\circ$) $\downarrow$}  & \multirow{2}{*}{\makecell{AUC\\($\%$) $\uparrow$}} \\
\cline{2-9} & Mean & Med. & Mean & Med. & Mean & Med. & Mean & Med. \\ 
\midrule
Upright~\cite{Lee:2014}                    &  3.43 & \underline{1.43} &  3.03 & \underline{1.13} &  6.85 & \textbf{0.47} &  8.62 & \underline{3.21} & 79.16 \\
A-Contrario~\cite{Simon:2018}              &  5.77 &  1.53 &  4.91 &  1.19 &  6.93 &  0.66 &   -   &   -   & 72.75 \\
DeepHorizon~\cite{Workman:2016}*           &  2.87 &  2.12 &  2.36 &  1.64 &  1.16 &  0.85 &   -   &   -   & \underline{80.65} \\
Perceptual~\cite{Hold-Geoffroy:2018}       & \underline{2.54} &  1.93 & \underline{2.11} &  1.49 & \underline{1.06} &  0.77 & \textbf{5.29} &  3.93 & \textbf{80.85} \\
Perceptual~\cite{Hold-Geoffroy:2018} $+\mathbf{L}$  &  2.86 &  2.17 &  2.45 &  1.76 & \underline{1.06} &  0.75  &  6.29 &  4.37 & 78.38 \\
UprightNet~\cite{Xian:2019}**              & 34.72 & 34.67 & 35.31 & 33.72 &  4.92 &  2.88 &   -   &   -   &   -   \\
\midrule
\textbf{Ours}                      & \textbf{2.33} &  \textbf{1.27} & \textbf{1.97} & \textbf{0.96} & \textbf{0.97} &  \underline{0.51} & \underline{5.66} & \textbf{3.16} & 80.07 \\
\bottomrule
\end{tabularx}
}
\end{table}

\subsection{Additional Results on Google Street View \cite{GSVI} Dataset}
\begin{figure*}[t!]
    \begin{center}
    \begin{tabular}{ccccc}
    \includegraphics[width=0.18\linewidth]{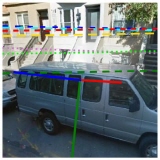} &
    \includegraphics[width=0.18\linewidth]{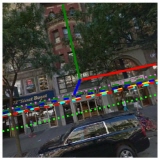} &
    \includegraphics[width=0.18\linewidth]{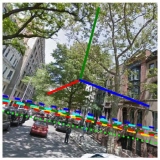} &
    \includegraphics[width=0.18\linewidth]{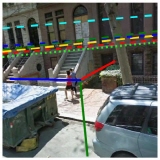} &
    \includegraphics[width=0.18\linewidth]{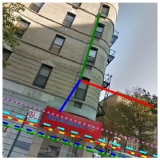} \\
    \includegraphics[width=0.18\linewidth]{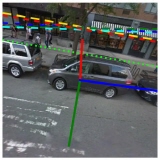} &
    \includegraphics[width=0.18\linewidth]{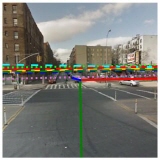} &
    \includegraphics[width=0.18\linewidth]{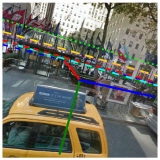} &
    \includegraphics[width=0.18\linewidth]{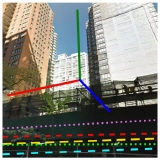} &
    \includegraphics[width=0.18\linewidth]{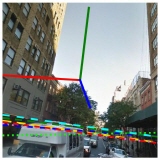} \\
    \includegraphics[width=0.18\linewidth]{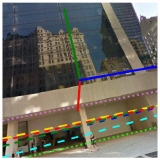} &
    \includegraphics[width=0.18\linewidth]{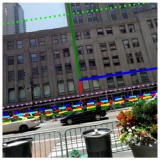} &
    \includegraphics[width=0.18\linewidth]{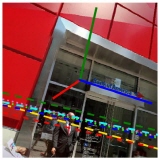} &
    \includegraphics[width=0.18\linewidth]{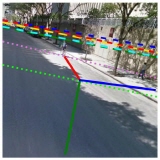} &
    \includegraphics[width=0.18\linewidth]{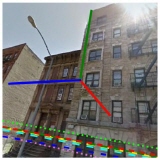} \\
    \includegraphics[width=0.18\linewidth]{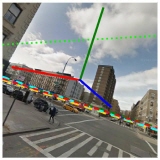} &
    \includegraphics[width=0.18\linewidth]{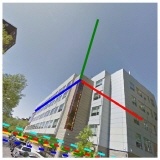} &
    \includegraphics[width=0.18\linewidth]{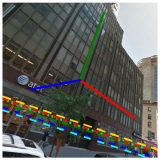} &
    \includegraphics[width=0.18\linewidth]{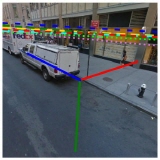} &
    \includegraphics[width=0.18\linewidth]{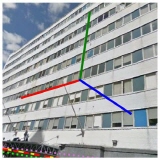} \\
    \includegraphics[width=0.18\linewidth]{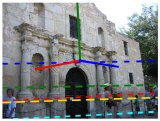} &
    \includegraphics[width=0.18\linewidth]{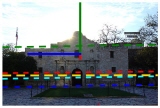} &
    \includegraphics[width=0.18\linewidth]{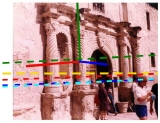} &
    \includegraphics[width=0.18\linewidth]{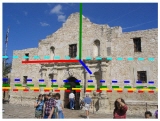} &
    \includegraphics[width=0.18\linewidth]{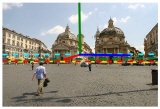} \\
    \includegraphics[width=0.18\linewidth]{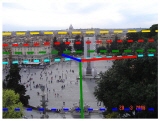} &
    \includegraphics[width=0.18\linewidth]{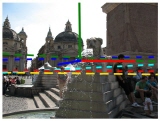} &
    \includegraphics[width=0.18\linewidth]{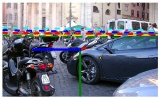} &
    \includegraphics[width=0.18\linewidth]{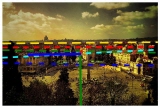} &
    \includegraphics[width=0.18\linewidth]{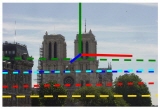} \\
    \includegraphics[width=0.18\linewidth]{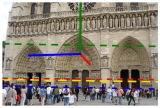} &
    \includegraphics[width=0.18\linewidth]{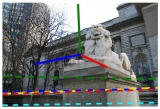} &
    \includegraphics[width=0.18\linewidth]{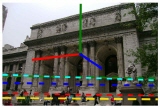} &
    \includegraphics[width=0.18\linewidth]{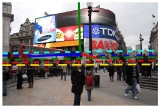} &
    \includegraphics[width=0.18\linewidth]{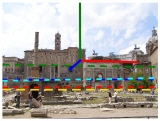} \\
    \includegraphics[width=0.18\linewidth]{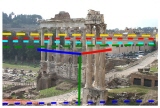} &
    \includegraphics[width=0.18\linewidth]{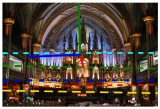} &
    \includegraphics[width=0.18\linewidth]{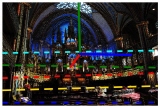} &
    \includegraphics[width=0.18\linewidth]{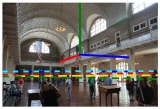} &
    \includegraphics[width=0.18\linewidth]{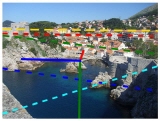} \\
    \end{tabular}
    \newcommand{\crule}[3][red]{\textcolor{#1}{\rule{#2}{#3} \rule{#2}{#3} \rule{#2}{#3} \rule{#2}{#3}}}
    {\scriptsize
    \begin{tabular}{llll}
    \crule[Goldenrod]{0.01\linewidth}{0.01\linewidth} Ground Truth & 
    \crule[Orchid]{0.01\linewidth}{0.01\linewidth} Upright \cite{Lee:2014} & 
    \crule[LimeGreen]{0.01\linewidth}{0.01\linewidth} A-Contario \cite{Simon:2018} & 
    \crule[blue]{0.01\linewidth}{0.01\linewidth} DeepHorizon \cite{Workman:2016} \\ 
    &
    \crule[OliveGreen]{0.01\linewidth}{0.01\linewidth} Perceptual \cite{Hold-Geoffroy:2018} & 
    \crule[SkyBlue]{0.01\linewidth}{0.01\linewidth} Perceptual \cite{Hold-Geoffroy:2018} $+ \mathbf{L}$  &  
    \crule[red]{0.01\linewidth}{0.01\linewidth} Ours \\
    \end{tabular}
    }
    \end{center}
    \caption{Examples of horizon line prediction on the Google Street View test set (top four rows) and on the HLW test set (bottom four rows). Each example also shows the Manhattan direction of the highest score candidate.}
    \label{fig:horizon_line_predictions_gsv}
\end{figure*}

\Fig{horizon_line_predictions_gsv} presents additional results on the Google Street View \cite{GSVI} dataset, as in ~\Fig{horizon_line_predictions}~\refpaper{}, visualizing horizon line predictions and weakly supervised Manhattan directions. In each example, we illustrate the Manhattan directions of the highest score candidate (\Eq{final_score}~\refpaper{}). In most cases, our method provides better horizon prediction results than those of previous state-of-the-art methods. Note that we only use the supervision of horizon lines and focal lengths (3DoF), yet we can further estimate the camera rotation and focal lengths (4DoF) based on the Manhattan world assumption.

\subsection{Visualization of Our Network I/Os}

\begin{figure*}[t!]
    \begin{center}
    \setlength\tabcolsep{0.5pt} 
    \begin{tabular}{ccccccc}
    \includegraphics[width=0.14\linewidth]{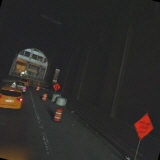} &
    \includegraphics[width=0.14\linewidth]{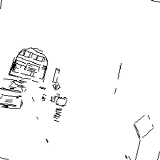} &
    \includegraphics[width=0.14\linewidth]{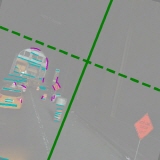} &
    \includegraphics[width=0.14\linewidth]{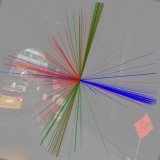} &
    \includegraphics[width=0.14\linewidth]{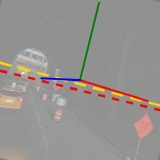} &
    \includegraphics[width=0.14\linewidth]{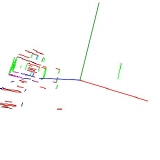} &
    \includegraphics[width=0.14\linewidth]{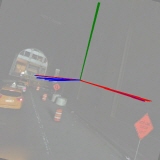} \\
    \includegraphics[width=0.14\linewidth]{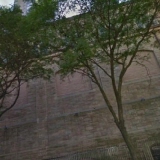} &
    \includegraphics[width=0.14\linewidth]{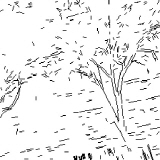} &
    \includegraphics[width=0.14\linewidth]{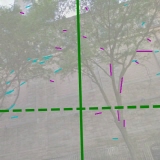} &
    \includegraphics[width=0.14\linewidth]{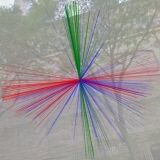} &
    \includegraphics[width=0.14\linewidth]{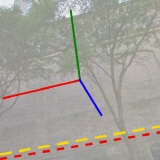} &
    \includegraphics[width=0.14\linewidth]{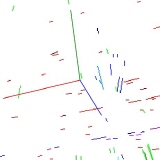} &
    \includegraphics[width=0.14\linewidth]{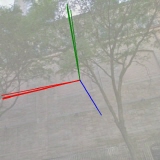} \\
    \includegraphics[width=0.14\linewidth]{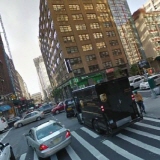} &
    \includegraphics[width=0.14\linewidth]{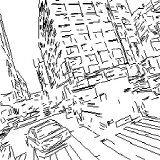} &
    \includegraphics[width=0.14\linewidth]{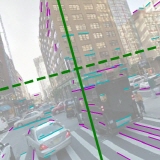} &
    \includegraphics[width=0.14\linewidth]{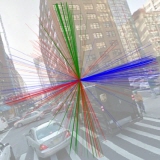} &
    \includegraphics[width=0.14\linewidth]{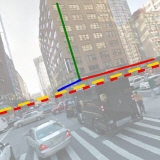} &
    \includegraphics[width=0.14\linewidth]{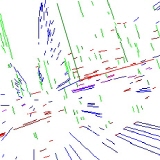} &
    \includegraphics[width=0.14\linewidth]{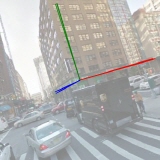} \\
    \includegraphics[width=0.14\linewidth]{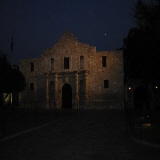} &
    \includegraphics[width=0.14\linewidth]{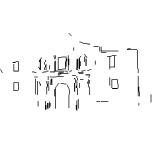} &
    \includegraphics[width=0.14\linewidth]{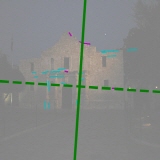} &
    \includegraphics[width=0.14\linewidth]{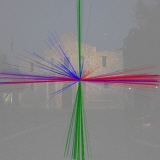} &
    \includegraphics[width=0.14\linewidth]{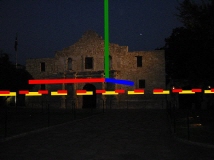} &
    \includegraphics[width=0.14\linewidth]{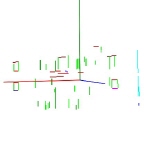} &
    \includegraphics[width=0.14\linewidth]{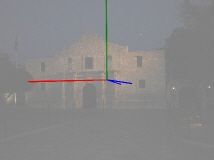} \\
    \includegraphics[width=0.14\linewidth]{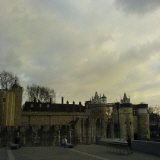} &
    \includegraphics[width=0.14\linewidth]{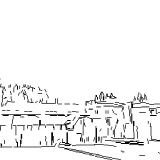} &
    \includegraphics[width=0.14\linewidth]{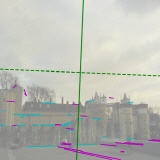} &
    \includegraphics[width=0.14\linewidth]{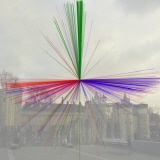} &
    \includegraphics[width=0.14\linewidth]{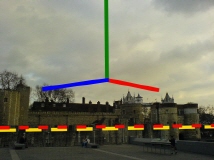} &
    \includegraphics[width=0.14\linewidth]{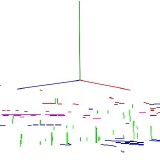} &
    \includegraphics[width=0.14\linewidth]{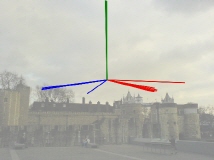} \\
    \includegraphics[width=0.14\linewidth]{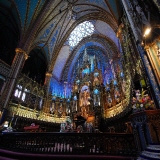} &
    \includegraphics[width=0.14\linewidth]{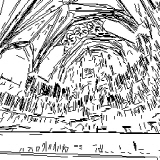} &
    \includegraphics[width=0.14\linewidth]{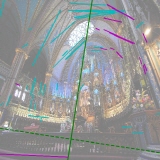} &
    \includegraphics[width=0.14\linewidth]{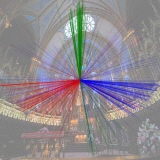} &
    \includegraphics[width=0.14\linewidth]{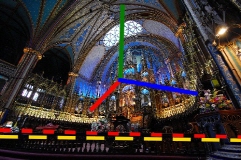} &
    \includegraphics[width=0.14\linewidth]{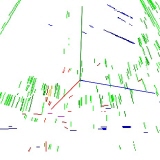} &
    \includegraphics[width=0.14\linewidth]{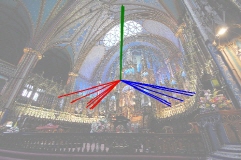} \\
    (a) & (b) & (c) & (d) & (e) & (f) & (g) \\
    \end{tabular}
    \end{center}
    \caption{Sampled images from the Google Street View test set (top three rows) and the HLW test set (bottom three rows).
    For each row, we show:
    (a) input image;
    (b) rasterized line segment map $\mathbf{L}$;
    (c) two groups of horizontal line segments (cyan \& magenta) used for sampling candidates of Manhattan directions;
    (d) sampled candidates of Manhattan directions;
    (e) ground truth and predicted horizon lines (yellow \& red dashed) as well as the estimated Manhattan directions of the highest score candidate;
    (f) activation map $\mathbf{A}$ of the Manhattan directions shown in (e); and
    (g) Manhattan directions of the top-8 high score candidates.}
    \label{fig:procedure}
\end{figure*}

\Fig{procedure} illustrates how geometric cues are processed and utilized in the proposed method. Each row of \Fig{procedure}(a)-(d) shows the input image, rasterized line segment map $\mathbf{L}$, grouped horizon line segments used for sampling candidates of Manhattan directions (as in Fig. 3(d) in the paper), and the set of sampled candidates of Manhattan directions, respectively.

\Fig{procedure}(e) displays the prediction results of the horizons and their corresponding ground truths, as well as the Manhattan direction of the highest score candidate. Activation maps $\mathbf{A}$ with respect to the Manhattan directions are presented in \Fig{procedure}(f). Notice that activation maps $\mathbf{A}$ in \Fig{procedure}(f) explain much of their respective line segment maps $\mathbf{L}$ in \Fig{procedure}(b), exemplifying how our method incorporates with the Manhattan world assumption. 

\Fig{procedure}(g) superimposes the eight Manhattan directions of the top-8 high-scoring candidates over the input images. As illustrated in \Fig{procedure}(g), the zenith directions are almost the same between candidates, as the man-made scenes usually satisfies either the Manhattan or Atlanta world assumption \cite{Coughlan:1999,Schindler:2004}. For scenes satisfying the Manhattan world assumption (rows 1--4), the axes of eight frames almost overlap. For the last two scenes (rows 5 and 6) that follow the Atlanta world assumption, all the frames have zenith directions that are very close to each other. By utilizing these frames we can robustly and accurately estimate horizon lines and focal lengths of given scenes. 

\subsection{Comparison of Manhattan Direction Prediction on YUD~\cite{YUD:2008} and ECD~\cite{ECD:2012}}

We report the accuracy of Manhattan direction prediction using our method on YUD~\cite{YUD:2008} and ECD~\cite{ECD:2012} datasets and compare the result with those of the other methods. In the experiment, we took the network model trained on the Google Street View dataset and tested it on YUD \cite{YUD:2008} and ECD \cite{ECD:2012} datasets. For the evaluation, the Manhattan direction of the high score candidate is used.

YUD \cite{YUD:2008} dataset contains 102 images under the Manhattan assumption, where each image is annotated with three VPs and a focal length. ECD \cite{ECD:2012} dataset contains 103 images under the Atlanta assumption, where each image is annotated with a zenith VP and more than two horizontal VPs on a horizon line. For ECD \cite{ECD:2012} dataset, the direction which is the closest to the prediction is used for comparisons.

\Tbl{comp_yud_ecd} shows the quantitative comparisons, in terms of the relative rotation angle, differences of FoV, and AUC. For FoV and AUC, we used the same setting as depicted in~\Sec{comparisons}~\refpaper{}. As shown~\Tbl{comp_yud_ecd}, our results are comparable to the ones of non-neural-net methods~\cite{Lee:2014,Simon:2018}, which are highly optimized for YUD~\cite{YUD:2008} and ECD~\cite{ECD:2012} datasets. We remark that our networks are trained on a different dataset and also with weak and indirect supervision (horizon lines and focal lengths).

\begin{table}[t!]
\renewcommand{\arraystretch}{0.9}
\caption{Quantitative evaluations of Manhattan direction prediction with YUD \cite{YUD:2008} and ECD \cite{ECD:2012} datasets.}
\label{tbl:comp_yud_ecd}
{\scriptsize
\begin{tabularx}{\textwidth}{c|c|CC|CC|C}
\toprule
\multirow{2}{*}{Dataset} & \multirow{2}{*}{Method} & \multicolumn{2}{c|}{Rotation Angle ($^\circ$) $\downarrow$}  &  \multicolumn{2}{c|}{FoV ($^\circ$) $\downarrow$} & \multirow{2}{*}{\makecell{AUC\\($\%$) $\uparrow$}} \\
\cline{3-6} & & Mean & Med. & Mean & Med. \\ 
\midrule
\multirow{3}{*}{YUD~\cite{YUD:2008}}
& Upright~\cite{Lee:2014}       &  \textbf{0.46} &  \textbf{0.27} & \underline{3.41} &  \underline{1.18} & \underline{87.26} \\
& A-Contrario~\cite{Simon:2018} &  1.03 &  0.78 & 3.42 &  \textbf{1.17} & \textbf{95.35} \\
& \textbf{Ours}                 &  \underline{0.52} & \underline{0.33} & \textbf{2.73} &  1.47 & 83.21 \\
\midrule
\multirow{3}{*}{ECD~\cite{ECD:2012}}
& Upright~\cite{Lee:2014}       & \textbf{2.91} & \textbf{1.02} & \textbf{12.44} & \textbf{6.31} & 76.71 \\
& A-Contrario~\cite{Simon:2018} & \underline{3.11} &  1.44 & 16.73 & 10.22 & \textbf{91.10} \\
& \textbf{Ours}                 &  3.14 & \underline{1.34} & \underline{13.61} & \underline{7.73} & \underline{77.61} \\
\bottomrule
\end{tabularx}
}
\end{table}


\subsection{Parameter Sensitivity Test}
\label{sec:parameter_sensitivity}

We tested parameter sensitivity by varying our parameters including: the angle threshold for vertical lines ($\delta_z$ in \Eq{angle_threshold}), the angle thresholds for deciding the positive and negative samples of zenith candidates ($\delta_p$ and $\delta_n$ in \Eq{zenith_label}), the score threshold for ZSNet ($\delta_c$ in \Eq{zenith_score}), the score threshold for FSNet ($\delta_s$ in \Eq{score_tol}), and the numbers of line segments and intersection points used in the network ($|L_z|$ and $|Z|$). Also, we tested sensitivity to line detection algorithm by varying the LSD algorithm parameter, $-\log{(\textrm{NFA})}$, where $\textrm{NFA}$ is the number of false alarms and also by replacing the line detection algorithm with MCMLSD \cite{Almazan:2017}. We used the network model trained with the Google Street View~\cite{GSVI} dataset with \emph{default} parameters and tested the model by varying the parameters, except for $\delta_p$ and $\delta_n$ in \Eq{zenith_label} and $\delta_s$ in \Eq{score_tol}; these parameters change either the ground truth labels or the loss function. For those parameters, we finetuned our network from the pretrained model. All results are reported in~\Tbl{parameter_sensitivity}. The highlighted rows show the results with default parameters. The results demonstrate that our method is robust to the change of the parameters. 

In our implementation, 1,024 for both lines and points was the maximum number to train the network with 11 GB GPU memory. However, more numbers of lines and points also significantly increase training time and GPU memory usage. For the sake of simplicity, all results reported in this paper were obtained with $|L_z|=|Z|=256$ both at training time and test time.

\begin{table}[t!]
\renewcommand{\arraystretch}{0.9}
\newcommand{\OursCellColor}{LightGoldenrod}
\caption{Parameter sensitivity test results. The highlighted rows show the result with the default parameters. Bold is the best result, and underscore is the second-best result in each experiment.}
\label{tbl:parameter_sensitivity}
{\scriptsize
\begin{tabularx}{\textwidth}{c|l|CC|CC|CC|CC|C}
\toprule
\multicolumn{2}{l|}{} & \multicolumn{2}{c|}{Angle ($^\circ$) $\downarrow$}  & \multicolumn{2}{c|}{Pitch ($^\circ$) $\downarrow$} & \multicolumn{2}{c|}{Roll ($^\circ$) $\downarrow$} & \multicolumn{2}{c|}{FoV ($^\circ$) $\downarrow$}  & \multirow{2}{*}{\makecell{AUC\\($\%$) $\uparrow$}} \\
\cline{3-10}
\multicolumn{2}{c|}{} & Mean & Med. & Mean & Med. & Mean & Med. & Mean & Med. & \\ 
\midrule
\multirow{3}{*}{\makecell{$\delta_z$ (\Eq{angle_threshold})}}
    & $58.5^\circ$ &  \textbf{2.04} &  \underline{1.67} &  \textbf{1.84} &  \underline{1.44} &  \textbf{0.64} &  \textbf{0.46} &  \textbf{5.67} &  \textbf{3.52} & \underline{83.01} \\
    &\cellcolor{\OursCellColor}$67.5^\circ$ &\cellcolor{\OursCellColor} \underline{2.12} &\cellcolor{\OursCellColor} \textbf{1.61} &\cellcolor{\OursCellColor} \underline{1.92} &\cellcolor{\OursCellColor} \textbf{1.38} &\cellcolor{\OursCellColor} \underline{0.75} &\cellcolor{\OursCellColor} \underline{0.47} &\cellcolor{\OursCellColor} \underline{6.01} &\cellcolor{\OursCellColor} \underline{3.72} &\cellcolor{\OursCellColor} \textbf{83.12} \\
    & $76.5^\circ$ &  2.83 &  1.97 &  2.30 &  1.67 &  1.62 &  0.57 &  6.47 &  3.97 & 79.70 \\
\midrule
\multirow{3}{*}{\makecell{$\delta_p$, $\delta_n$ (\Eq{zenith_label})}}
    & $1^\circ$, $2^\circ$ &  3.45 &  1.98 &  2.73 &  1.87 &  1.53 &  0.62 &  7.43 &  \underline{4.02} & 75.22 \\
    &\cellcolor{\OursCellColor}$2^\circ$, $5^\circ$ &\cellcolor{\OursCellColor} \textbf{2.12} &\cellcolor{\OursCellColor} \textbf{1.61} &\cellcolor{\OursCellColor} \textbf{1.92} &\cellcolor{\OursCellColor} \textbf{1.38} &\cellcolor{\OursCellColor} \textbf{0.75} &\cellcolor{\OursCellColor} \textbf{0.47} &\cellcolor{\OursCellColor} \textbf{6.01} &\cellcolor{\OursCellColor} \textbf{3.72} &\cellcolor{\OursCellColor} \textbf{83.12} \\
    & $5^\circ$, $10^\circ$ &  \underline{2.54} &  \underline{1.97} &  \underline{2.10} &  \underline{1.71} &  \textbf{0.75} &  \underline{0.57} &  \underline{6.64} &  4.21 & \underline{79.01} \\
\midrule
\multirow{3}{*}{\makecell{$\delta_c$ (\Eq{zenith_score})}}
    & 0.4  & 2.32 &  1.84 &  2.28 &  \underline{1.44} &  0.84 &  0.52 &  6.82 &  4.37 & 80.23 \\
    &\cellcolor{\OursCellColor}0.5 &\cellcolor{\OursCellColor} \textbf{2.12} &\cellcolor{\OursCellColor} \textbf{1.61} &\cellcolor{\OursCellColor} \textbf{1.92} &\cellcolor{\OursCellColor} \textbf{1.38} &\cellcolor{\OursCellColor} \underline{0.75} &\cellcolor{\OursCellColor} \underline{0.47} &\cellcolor{\OursCellColor} \underline{6.01} &\cellcolor{\OursCellColor} \underline{3.72} &\cellcolor{\OursCellColor} \textbf{83.12} \\
    & 0.6 & \underline{2.17} & \underline{1.71} &  \underline{1.96} &  1.48 &  \textbf{0.65} &  \textbf{0.46} &  \textbf{5.76} &  \textbf{3.42} & \underline{82.94} \\
\midrule
\multirow{3}{*}{\makecell{$\delta_s$ (\Eq{score_tol})}}
    & 0.4  &  3.02 &  1.80 &  2.71 &  1.61 &  1.04 &  \textbf{0.47} &  6.70 &  4.05 & 80.82 \\
    &\cellcolor{\OursCellColor}0.5 &\cellcolor{\OursCellColor} \textbf{2.12} &\cellcolor{\OursCellColor} \textbf{1.61} &\cellcolor{\OursCellColor} \textbf{1.92} &\cellcolor{\OursCellColor} \textbf{1.38} &\cellcolor{\OursCellColor} \underline{0.75} &\cellcolor{\OursCellColor} \textbf{0.47} &\cellcolor{\OursCellColor} \underline{6.01} &\cellcolor{\OursCellColor} \underline{3.72} &\cellcolor{\OursCellColor} \textbf{83.12} \\
    & 0.6 & \underline{2.19} & \underline{1.64} & \underline{1.93} & \underline{1.43} & \textbf{0.74} & \textbf{0.47} & \textbf{5.88} & \textbf{3.43} & \underline{83.10} \\
\midrule
\multirow{4}{*}{\makecell{top-$k$}}
    & $k=1$  &  2.23 &  1.72 &  1.97 &  1.49 &  0.75 &  0.55 &  6.71 &  3.84 & 82.12 \\
    & $k=4$  &  \textbf{2.10} & \underline{1.70} &  \textbf{1.89} & \underline{1.48} & \textbf{0.65} &  0.49 & \underline{6.01} & \textbf{3.66} & \underline{83.05} \\
    &\cellcolor{\OursCellColor}$k=8$ &\cellcolor{\OursCellColor} \underline{2.12} &\cellcolor{\OursCellColor} \textbf{1.61} &\cellcolor{\OursCellColor} \underline{1.92} &\cellcolor{\OursCellColor} \textbf{1.38} &\cellcolor{\OursCellColor} 0.75 &\cellcolor{\OursCellColor} \underline{0.47} &\cellcolor{\OursCellColor} \underline{6.01} &\cellcolor{\OursCellColor} 3.72 &\cellcolor{\OursCellColor} \textbf{83.12} \\
    & $k=16$ &  2.24 &  1.71 &  2.04 &  1.52 & \textbf{0.65} & \textbf{0.46} &  \textbf{5.61} & \textbf{3.66} & 82.70 \\
\midrule
\multirow{2}{*}{\makecell{$|L_z|$, $|Z|$}}
    & \cellcolor{\OursCellColor}256, 256 &\cellcolor{\OursCellColor} \underline{2.12} &\cellcolor{\OursCellColor} \textbf{1.61} &\cellcolor{\OursCellColor} \underline{1.92} &\cellcolor{\OursCellColor} \textbf{1.38} &\cellcolor{\OursCellColor} \underline{0.75} &\cellcolor{\OursCellColor} \underline{0.47} &\cellcolor{\OursCellColor} \underline{6.01} &\cellcolor{\OursCellColor} \underline{3.72} &\cellcolor{\OursCellColor} \underline{83.12} \\
    & 1024, 1024 & \textbf{2.05} &  \underline{1.65} &  \textbf{1.86} &  \underline{1.46} & \textbf{0.63} & \textbf{0.45} &  \textbf{5.66} & \textbf{3.45} & \textbf{83.80} \\
\midrule
\multirow{6}{*}{\makecell{$-\log{(\textrm{NFA})}$\\in LSD~\cite{Gioi:2010},\\MCMLSD~\cite{Almazan:2017}}}
    &\cellcolor{\OursCellColor}0   &\cellcolor{\OursCellColor} \underline{2.12} &\cellcolor{\OursCellColor} \textbf{1.61} &\cellcolor{\OursCellColor} 2.09 &\cellcolor{\OursCellColor} \textbf{1.38} &\cellcolor{\OursCellColor} 0.80 &\cellcolor{\OursCellColor} \underline{0.47} &\cellcolor{\OursCellColor} 6.15 &\cellcolor{\OursCellColor} 3.72 &\cellcolor{\OursCellColor} 83.12 \\
    &$0.01\times1.75^{0}$       & \underline{2.12} & 1.74 & \textbf{1.91} & 1.54 & \textbf{0.65} & \underline{0.47} & 6.02 & 3.77 & 82.38 \\
    &$0.01\times1.75^{5}$       & \textbf{2.11} & 1.72 & \textbf{1.91} & 1.51 & \textbf{0.65} & 0.48 & 6.07 & 3.90 & \textbf{83.36} \\
    &$0.01\times1.75^{10}$      & 2.19 & 1.75 & 1.95 & 1.55 & 0.71 & 0.49 & 6.25 & 3.65 & 82.97 \\
    &$0.01\times1.75^{15}$      & 2.17 & 1.70 & 1.95 & \underline{1.46} & 0.67 & \textbf{0.46} & \textbf{5.53} & \underline{3.16} & \underline{83.34} \\
    &MCMLSD~\cite{Almazan:2017} & 2.31 & \underline{1.65} & 2.02 & \underline{1.46} & 0.81 & 0.50 & \underline{5.85} & \textbf{3.01} & 83.05 \\
\bottomrule
\end{tabularx}
}
\end{table}

\subsection{Visualization of FSNet Focus}

\begin{figure*}[t!]
    \begin{center}
    \begin{tabular}{cccccc}
    \includegraphics[width=0.15\linewidth]{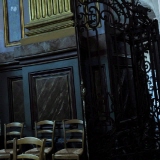} &
    \includegraphics[width=0.15\linewidth]{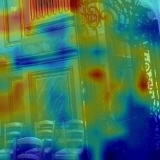} &
    \includegraphics[width=0.15\linewidth]{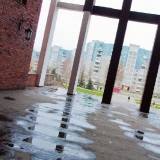} &
    \includegraphics[width=0.15\linewidth]{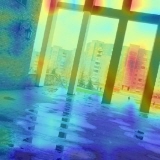} &
    \includegraphics[width=0.15\linewidth]{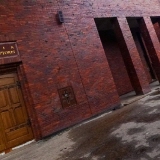} &
    \includegraphics[width=0.15\linewidth]{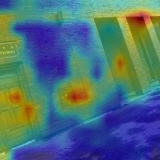} \\
    \includegraphics[width=0.15\linewidth]{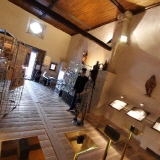} &
    \includegraphics[width=0.15\linewidth]{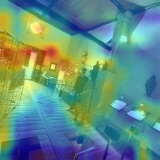} &
    \includegraphics[width=0.15\linewidth]{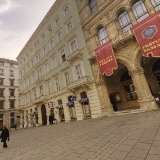} &
    \includegraphics[width=0.15\linewidth]{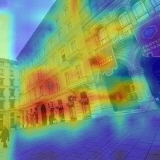} &
    \includegraphics[width=0.15\linewidth]{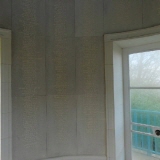} &
    \includegraphics[width=0.15\linewidth]{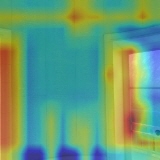} \\
    \includegraphics[width=0.15\linewidth]{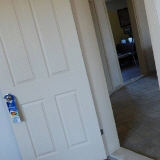} &
    \includegraphics[width=0.15\linewidth]{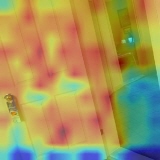} &
    \includegraphics[width=0.15\linewidth]{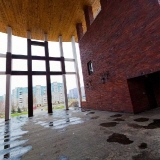} &
    \includegraphics[width=0.15\linewidth]{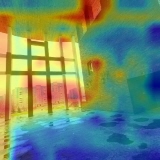} &
    \includegraphics[width=0.15\linewidth]{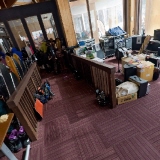} &
    \includegraphics[width=0.15\linewidth]{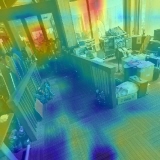} \\
    \includegraphics[width=0.15\linewidth]{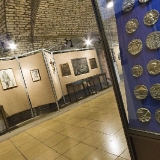} &
    \includegraphics[width=0.15\linewidth]{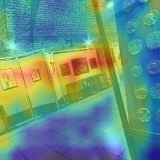} &
    \includegraphics[width=0.15\linewidth]{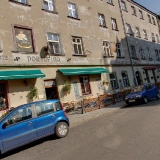} &
    \includegraphics[width=0.15\linewidth]{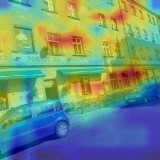} &
    \includegraphics[width=0.15\linewidth]{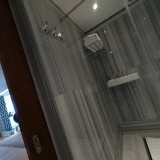} &
    \includegraphics[width=0.15\linewidth]{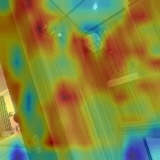} \\
    \includegraphics[width=0.15\linewidth]{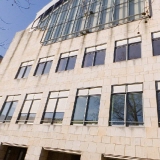} &
    \includegraphics[width=0.15\linewidth]{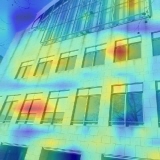} &
    \includegraphics[width=0.15\linewidth]{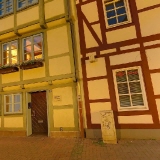} &
    \includegraphics[width=0.15\linewidth]{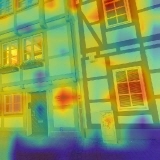} &
    \includegraphics[width=0.15\linewidth]{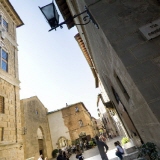} &
    \includegraphics[width=0.15\linewidth]{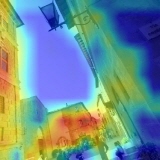} \\
    \includegraphics[width=0.15\linewidth]{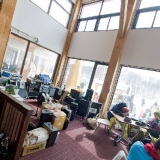} &
    \includegraphics[width=0.15\linewidth]{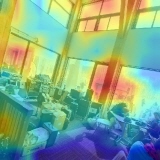} &
    \includegraphics[width=0.15\linewidth]{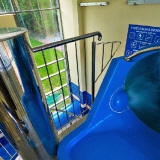} &
    \includegraphics[width=0.15\linewidth]{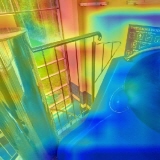} &
    \includegraphics[width=0.15\linewidth]{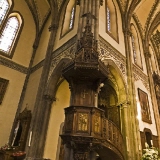} &
    \includegraphics[width=0.15\linewidth]{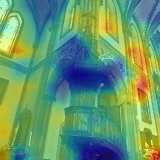} \\
    \includegraphics[width=0.15\linewidth]{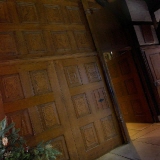} &
    \includegraphics[width=0.15\linewidth]{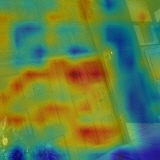} &
    \includegraphics[width=0.15\linewidth]{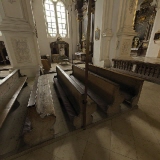} &
    \includegraphics[width=0.15\linewidth]{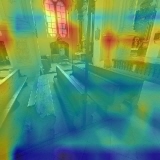} &
    \includegraphics[width=0.15\linewidth]{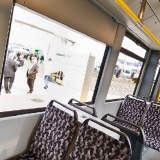} &
    \includegraphics[width=0.15\linewidth]{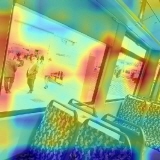} \\
    \end{tabular}
    \end{center}
    \caption{More visualizations of FSNet focus: Left is input; right is feature highlight.}
    \label{fig:more_feature_highligh}
\end{figure*}

In \Fig{more_feature_highligh}, we show more visualizations of the weights of the second last convolution layer in FSNet, as shown in~\Fig{feature_highlight}~\refpaper{}. The network mostly focuses on the lines that pass the vanishing points.

\subsection{Failure Cases}

\begin{figure*}[t!]
    \begin{center}
    \begin{tabular}{ccccc}
    \includegraphics[width=0.18\linewidth]{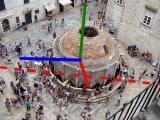} &
    \includegraphics[width=0.18\linewidth]{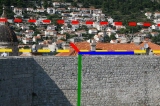} &
    \includegraphics[width=0.18\linewidth]{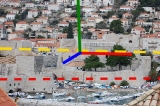} &
    \includegraphics[width=0.18\linewidth]{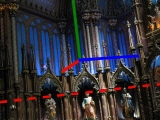} &
    \includegraphics[width=0.18\linewidth]{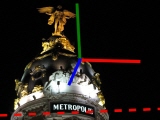} \\
    \includegraphics[width=0.18\linewidth]{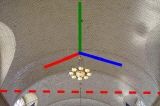} &
    \includegraphics[width=0.18\linewidth]{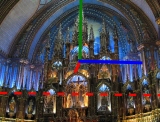} &
    \includegraphics[width=0.18\linewidth]{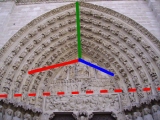} &
    \includegraphics[width=0.18\linewidth]{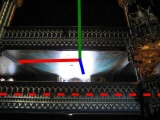} &
    \includegraphics[width=0.18\linewidth]{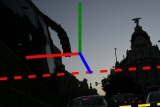} \\
    \includegraphics[width=0.18\linewidth]{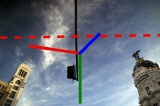} &
    \includegraphics[width=0.18\linewidth]{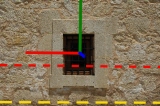} &
    \includegraphics[width=0.18\linewidth]{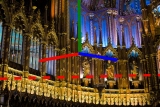} &
    \includegraphics[width=0.18\linewidth]{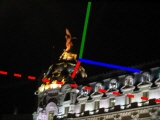} &
    \includegraphics[width=0.18\linewidth]{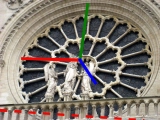} \\
    \end{tabular}
    \newcommand{\crule}[3][red]{\textcolor{#1}{\rule{#2}{#3} \rule{#2}{#3} \rule{#2}{#3} \rule{#2}{#3}}}
    {\scriptsize
    \begin{tabular}{ll}
    \crule[Goldenrod]{0.01\linewidth}{0.01\linewidth} Ground Truth & 
    \crule[red]{0.01\linewidth}{0.01\linewidth} Ours \\
    \end{tabular}
    }
    \end{center}
    \caption{Failure cases.}
    \label{fig:failure}
\end{figure*}

\Fig{failure} shows failure cases of our framework. The failure cases occur when the computation of the focal length is unstable, such that: i) the scene is far from the Manhattan assumption, ii) only short or noisy line segments are detected in the scene, iii) the scene is almost perpendicular to the center of projection. Please notice that the estimated zenith directions are still reasonable in \Fig{failure}, thanks to the semantic information learned by ResNet, the backbone of our FSNet. Therefore, even in the cases of \Fig{failure}, our framework is still applicable to image rotation corrections as shown in \Fig{teaser}(a).

\subsection{Experiment on KITTI~\cite{KITTI:2012} Dataset}
\label{sec:test_kitti}

We conducted an additional experiment with KITTI~\cite{KITTI:2012} dataset. The KITTI dataset contains wide-images captured by driving around urban cities and rural areas. We sample 8,675 images of urban scenes from the KITTI dataset and feed them to finetune our network from the pretrained model with the Google Street View dataset. We test our finetuned model to 481 images of urban and rural scenes from the KITTI dataset.

\begin{figure*}[t!]
    \begin{center}
    \begin{tabular}{ccc}
    \includegraphics[width=0.30\linewidth]{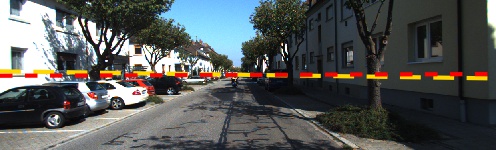} &
    \includegraphics[width=0.30\linewidth]{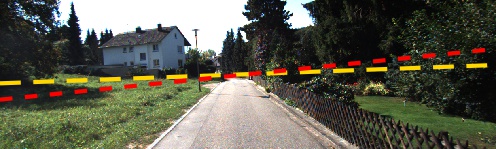} &
    \includegraphics[width=0.30\linewidth]{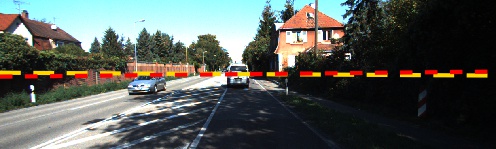} \\
    \includegraphics[width=0.30\linewidth]{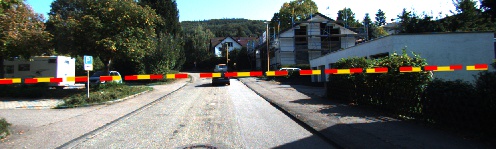} &
    \includegraphics[width=0.30\linewidth]{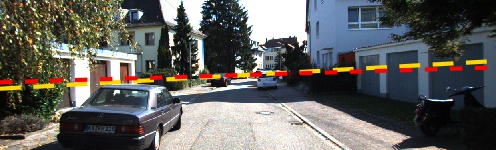} &
    \includegraphics[width=0.30\linewidth]{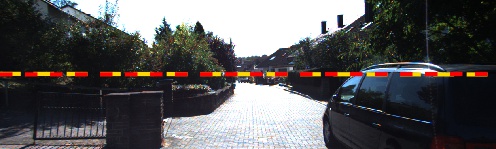} \\
    \includegraphics[width=0.30\linewidth]{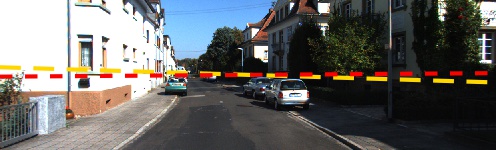} &
    \includegraphics[width=0.30\linewidth]{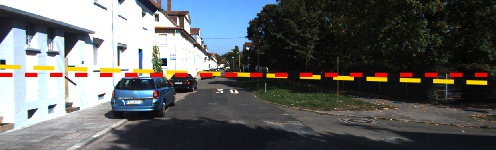} &
    \includegraphics[width=0.30\linewidth]{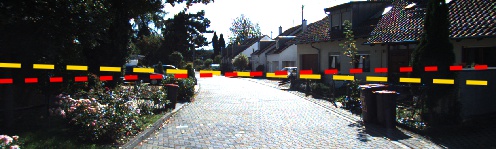} \\
    \includegraphics[width=0.30\linewidth]{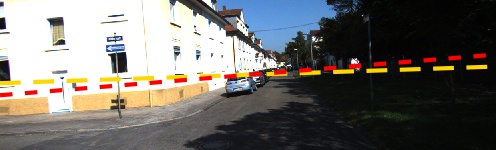} &
    \includegraphics[width=0.30\linewidth]{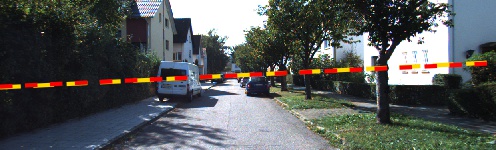} &
    \includegraphics[width=0.30\linewidth]{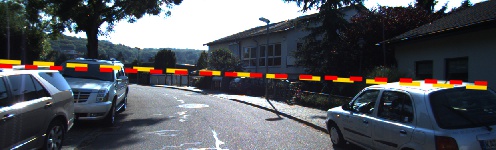} \\
    \includegraphics[width=0.30\linewidth]{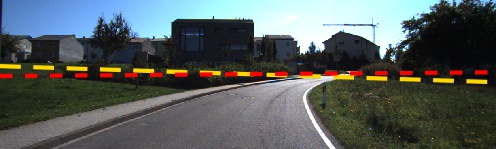} &
    \includegraphics[width=0.30\linewidth]{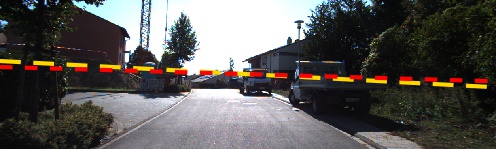} &
    \includegraphics[width=0.30\linewidth]{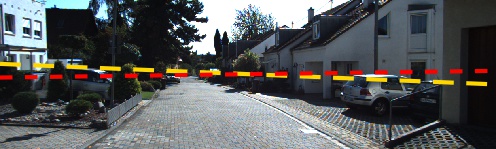} \\
    \includegraphics[width=0.30\linewidth]{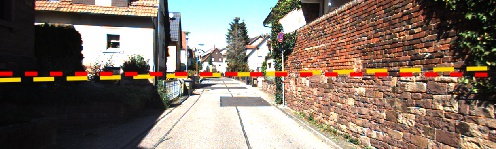} &
    \includegraphics[width=0.30\linewidth]{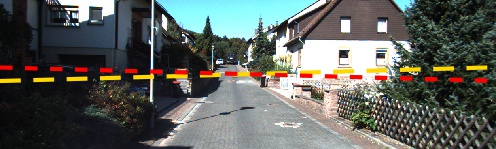} &
    \includegraphics[width=0.30\linewidth]{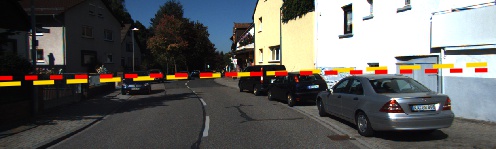} \\
    \includegraphics[width=0.30\linewidth]{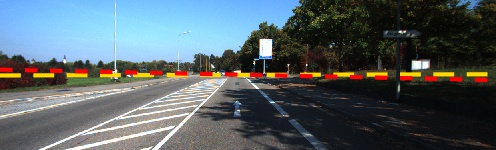} &
    \includegraphics[width=0.30\linewidth]{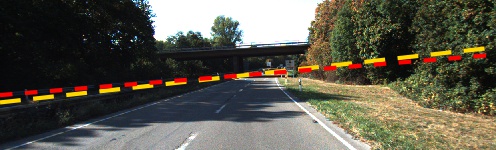} &
    \includegraphics[width=0.30\linewidth]{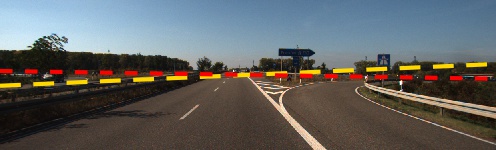} \\
    \includegraphics[width=0.30\linewidth]{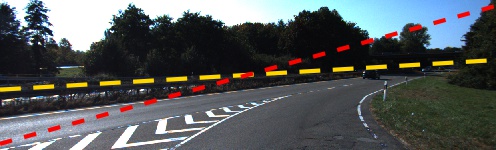} &
    \includegraphics[width=0.30\linewidth]{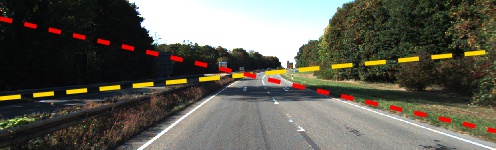} &
    \includegraphics[width=0.30\linewidth]{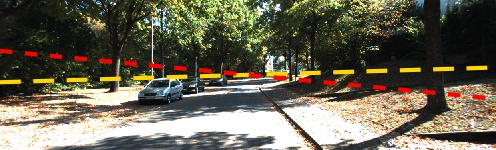} \\
    \end{tabular}
    \newcommand{\crule}[3][red]{\textcolor{#1}{\rule{#2}{#3} \rule{#2}{#3} \rule{#2}{#3} \rule{#2}{#3}}}
    {\scriptsize
    \begin{tabular}{ll}
    \crule[Goldenrod]{0.01\linewidth}{0.01\linewidth} Ground Truth & 
    \crule[red]{0.01\linewidth}{0.01\linewidth} Ours \\
    \end{tabular}
    }
    \end{center}
    \caption{Examples of horizon line prediction on the KITTI test set.}
    \label{fig:kitti}
\end{figure*}

\Fig{kitti} shows some examples of horizon predictions with the KITTI test set. Unfortunately, the GT horizons of the dataset are geometrically inaccurate due to the large influence of the vehicle's tilting angle during cornering. Nevertheless, we obtained interesting results where the estimated horizons of our framework do not deviate significantly from the GT horizons in urban areas. We believe the results come from the KITTI dataset, since there are little changes in the horizontal line and focal length. 
Another reason seems to be that the ResNet, the backbone of our FSNet, learned the scene context from the KITTI dataset. \Tbl{test_kitti}~reports the quantitative evaluations with KITTI dataset.

\begin{table}[t!]
\renewcommand{\arraystretch}{0.9}
\caption{Quantitative evaluations with KITTI dataset.}
\vspace{-0.5\baselineskip}
\label{tbl:test_kitti}
{\scriptsize
\begin{tabularx}{\textwidth}{l|CC|CC|CC|CC|C}
\toprule
\multirow{2}{*}{Method} & \multicolumn{2}{c|}{Angle ($^\circ$) $\downarrow$}  & \multicolumn{2}{c|}{Pitch ($^\circ$) $\downarrow$} & \multicolumn{2}{c|}{Roll ($^\circ$) $\downarrow$} & \multicolumn{2}{c|}{FoV ($^\circ$) $\downarrow$}  & \multirow{2}{*}{\makecell{AUC\\($\%$) $\uparrow$}} \\
\cline{2-9} & Mean & Med. & Mean & Med. & Mean & Med. & Mean & Med. \\ 
\midrule
\textbf{Ours ($k=8$)} & 3.38 & 3.06 & 2.50 & 2.12 & 1.87 & 1.88 & 17.68 & 16.49 & 78.34 \\
\bottomrule
\end{tabularx}
}
\end{table}